%% file: main.tex
\documentclass{article}

\usepackage[T1]{fontenc}
\usepackage[utf8]{inputenc}
\usepackage[margin=1in]{geometry}
\usepackage{graphicx}
\usepackage{tikz}
\usepackage{float}
\usepackage{subcaption}
\usepackage{amsmath}
\usepackage{amssymb}
\usepackage{booktabs}
\usepackage{multirow}
\usepackage{xcolor}
\usepackage[round]{natbib}
\usepackage[colorlinks=true,linkcolor=blue,citecolor=blue,urlcolor=blue]{hyperref}

\newcommand{\SELECTIVE}{\textsc{selective}}
\newcommand{\CONSTANT}{\textsc{constant}}
\newcommand{\DECODABLE}{\textsc{decodable}}

\newcommand{\GELU}{\ifmmode\operatorname{GELU}\else\textsc{gelu}\fi}
\DeclareMathOperator{\relu}{ReLU}
\newcommand{\sig}{\sigma}

\title{Training, Reading, and Editing Legible Transformers}
\author{Mark Oskin\\
Professor\\
School of Computer Science and Engineering\\
University of Washington\\
\texttt{mhoskin@uw.edu}}
\date{July 2026}

\begin{document}

\maketitle

\input{abstract}

\input{intro}
\input{background}
\input{architecture}
\input{train}
\input{understand}
\input{edit}
\input{dial}
\input{discussion}
\input{conclusion}

\appendix
\input{appendix}

\bibliographystyle{plainnat}
\bibliography{refs}

\end{document}

%% file: abstract.tex
\begin{abstract}
A transformer can be built from operators that are legible by construction --- bounded, named units that read as fuzzy set operations rather than dense activations --- but legibility must be pressed for during training, and the pressure has a failure mode. A crispness penalty meant to sharpen a bounded operator into a decisive detector instead collapses it into a dead constant. An identity, $\mathbb{E}[v(1-v)]=\mu(1-\mu)-\mathrm{var}$, shows why --- the penalty is a variance-minimizer blind to the difference between a live detector and a constant --- and names the fix: a per-channel variance floor, the target legibility metric written as a loss, which recovers both legibility and quality. A learned per-unit fraction then retires the hand-set reserved-$\GELU$ partition of prior work: given the choice the model keeps \emph{no} unit as pure $\GELU$ and routes $87\%$ of its load-bearing computation through crisp operators. The result is the most legible transformer we have built --- $78\%$ of its feed-forward operands and $50\%$ of its attention value channels are crisp-and-contextual detectors, and per-head legibility rises from $18\%$ in shallow layers to $78\%$ in deep ones. Read in the correct rotated per-layer frame, these units separate a clean \emph{detection} (what a unit responds to) from a harder \emph{naming} (what its output decodes to); and because the objective makes each unit crisp and sparse, edits to them are far more local --- $50$--$184\times$ in the deep layers where the edit sites concentrate --- and can target explicit conjunctions a single neuron cannot express. Finally, a \emph{between}-unit decorrelation pressure exposes a legibility dial: it trades a circuit's reuse for independence at no quality cost, turning concepts into single, surgically editable units and a prediction into a short explanation read off a handful of named operations. Quality holds at parity with a conventional baseline throughout.
\end{abstract}

%% file: intro.tex
\section{Introduction}
\label{sec:intro}

Most interpretability reconstructs meaning after training: a dense activation is probed, decomposed with a sparse autoencoder, or decoded through the vocabulary, and a human names what was found. An alternative is to build the meaning in --- to make a model's operators legible by construction, so that a coordinate carries a fixed, stated reading rather than one recovered post hoc. Two prior constructions take this route: the feed-forward layer becomes fuzzy set operations on bounded operands \citep{oskin2026boolglu}, and the attention value becomes bounded, named detectors \citep{oskin2026attention}, each at language-model parity (Figure~\ref{fig:arch}). But a legible-by-construction \emph{substrate} is not yet a legible \emph{model}. Its operators read cleanly only if training keeps them so; the two constructions have never been trained together; and reading such a model, and editing it, are problems of their own. This paper is about making that line usable end to end --- how to train the model so its units are actually legible, how to read them, how to edit them, and how a single further pressure reshapes the circuits they form.

The obstacle is the very pressure legibility depends on. A bounded operator reads cleanly only when it fires \emph{decisively} --- near $0$ or near $1$, ``absent'' or ``present'' --- so both constructions add a crispness penalty that pushes each value to a rail. Pushed too hard, that penalty does the opposite of its purpose: it drives a unit to a fixed value that ignores its input, a \emph{dead constant}, perfectly crisp and carrying no information, and quality collapses with it. The cause is exact. For a bounded value the crispness penalty equals $\mu(1-\mu)-\mathrm{var}$, a variance-minimizer that cannot tell a live contextual detector from a dead constant, so gradient descent takes the cheaper constant. Naming the cause gives the cure: a per-channel \emph{variance floor} --- the property that separates a live detector from a dead one, written directly as a loss. It adds one term and no parameters, and turns a pressure that ruins the model into one that recovers both legibility and quality. The same discipline retires a second hand-set dial. Prior work reserved a fixed fraction of feed-forward units as ordinary $\GELU$ for trainability; a per-unit learned gate turns that reservation into a decision the model makes, and given the choice it keeps \emph{no} unit as pure $\GELU$ and routes $87\%$ of its load-bearing computation through the legible operators. The result is the most legible transformer we have built, and the first legible on both sublayers at once: $78\%$ of its feed-forward operands and $50\%$ of its attention value channels are crisp, contextual detectors.

A legible model can be read, but reading it well takes care. A unit's write is illegible through the raw vocabulary projection in the middle of the network, where the residual sits in a rotated frame; a per-layer tuned lens rotates it back, and the abstract computation behind a completion becomes visible several layers earlier than the raw lens shows. Reading in that frame separates two properties the construction lets us measure apart: a unit's \emph{detection} --- the clean, input-dependent bit it responds to --- from its \emph{naming} --- whether what it writes decodes to a word. The model detects far more cleanly than it names: most feed-forward operands are selective detectors, but only about half decode to a vocabulary category, and the concepts it cannot name are the abstract ones it carries non-lexically through mid-stack. Those are also the concepts it \emph{distributes}: token-level features live on single units, while abstract concepts are spread across many and grow more distributed with depth --- the superposition of ordinary networks, now visible against a legible backdrop.

Because the objective makes each unit crisp and sparse, the standard tools for editing a transformer --- difference-of-means writes, activation steering --- apply to it better. A crisp, sparse unit is a clean site: an edit placed there stays local, $184\times$ more local in the deep layers where the editable sites concentrate, and needs no search to find. And the named set operations make possible an edit no single neuron can express --- a conjunction $A\cap B$ written at one addressable, two-operand unit. Editing borrows its mechanism from prior work; what the objective supplies is the quality of the site.

Reading the model's circuits shows how it wires a concept: redundantly, a few operators broadcast to many readers --- a \emph{fan-out} circuit, efficient but hard to touch, since editing one concept means altering every copy and disturbing every reader. A single further pressure reshapes it. A \emph{between}-unit decorrelation penalty --- forbidding units to co-fire, and so forbidding the shared building blocks that reuse depends on --- trades the fan-out circuit for a \emph{fan-in} one: many independent units, each aggregating a wide, disjoint input, at no cost to quality. The trade buys editability directly --- a concept that lived in a $562$-unit tangle becomes one addressable unit --- and legibility of a whole prediction: a token's attribution, spread across a thousand small pieces in an ordinary transformer, collapses onto a short list of named operations one can read as a sentence (\emph{``than'' because a comparative unit fired and a head read the comparative adjective}). What it spends is the compression reuse provides, a cost on out-of-distribution composition we do not yet measure.

Through all of this the through-line is that the contribution is \emph{legibility}, not a different machine or a better score. The structure these tools read off --- where a concept lives, how redundantly it is built, how widely it is read --- is structure a conventional transformer plausibly shares, and every measurement is defined on activations and weights a $\GELU$ model has too; what the construction adds is that all of it can be read directly, without a post-hoc dictionary that must be trained, chosen, and re-validated. Quality holds at parity with a conventional baseline throughout.

\medskip
\noindent\textbf{Contributions.}
\begin{itemize}
\item \textbf{A legibility objective.} A per-channel variance floor --- the legibility metric written as a loss --- repairs the collapse the crispness penalty otherwise causes, recovering both legibility and quality; the $\mu(1-\mu)-\mathrm{var}$ identity names why the naive penalty fails (Section~\ref{sec:train}).
\item \textbf{An end-to-end legible transformer.} The feed-forward and attention constructions are trained together for the first time, and a learned per-unit gate retires the hand-set reserved-$\GELU$ partition, yielding a model legible on \emph{both} sublayers at parity (Sections~\ref{sec:architecture}--\ref{sec:train}).
\item \textbf{Reading the units.} Read in each layer's rotated frame, the units separate a clean \emph{detection} from a harder \emph{naming}; the model detects far more than it names, and distributes exactly the concepts it cannot name (Section~\ref{sec:understand}).
\item \textbf{Editing the units.} Crisp, sparse units are surgical edit sites --- far more local than a conventional neuron --- and support a single-unit conjunctive $A\cap B$ edit no neuron can express (Section~\ref{sec:edit}).
\item \textbf{A legibility dial.} A between-unit decorrelation pressure trades a circuit's reuse (fan-out) for independence (fan-in) at no quality cost, making concepts surgically editable and a whole prediction explainable as a short list of named operations (Section~\ref{sec:dial}).
\end{itemize}

%% file: background.tex
\section{Background and Related Work}
\label{sec:background}

Our work sits at the intersection of five lines: interpretable-by-construction model components, training objectives for interpretability, vocabulary-space readout of hidden states, superposition and its consequences, and model editing. We build directly on prior results in each.

\subsection{Interpretable-by-construction components}
\label{sec:bg-construction}
A body of work replaces a standard component with one whose computation is legible by design, rather than recovering structure after training. Softmax Linear Units raise the fraction of first-glance-interpretable MLP neurons at little cost, while noting that such changes can also hide features \citep{elhage2022solu}. Bilinear MLPs remove the elementwise nonlinearity so that interaction structure is recoverable from the weights alone, at language-model parity \citep{pearce2025bilinear}. Codebook Features quantize the residual stream into a small set of discrete codes \citep{tamkin2024codebook}, and Backpack models express each token as a nonnegative combination of linearly decodable sense vectors, enabling predictable intervention at GPT-2 parity \citep{hewitt2023backpack}. Kolmogorov--Arnold Networks expose learned univariate functions on edges \citep{liu2024kan}. Our multiplicative operators are, mechanically, gated bilinear units in the GLU family \citep{shazeer2020glu}; what is specific is that both operands are sigmoid-bounded to $[0,1]$ and read as fuzzy set memberships, with an explicit set-difference term as a first-class unit.

A parallel neuro-symbolic literature realizes logical connectives as differentiable layers: Logic Tensor Networks \citep{badreddine2022ltn}, Logical Neural Networks \citep{riegel2020lnn}, the analysis of fuzzy t-norms under gradient descent \citep{vankrieken2022fuzzy}, logic-structured language representations \citep{chen2023folnet}, tensor logic \citep{domingos2025tensorlogic}, and interpretable logic classifiers \citep{perreault2025neurallogic}. These ground \emph{externally specified} formulae; our units instead carry emergent structure read out of a model trained only on next-token prediction. Post-training Boolean-operator calculi over frozen embeddings \citep{vexler2026nsfl} share the vocabulary but are not architectural. The fuzzy-Boolean feed-forward units and bounded value heads themselves are introduced in our prior work \citep{oskin2026boolglu,oskin2026attention}; Section~\ref{sec:architecture} recaps that construction so the present paper is self-contained. This paper's contribution is not the operators but the objective that makes them crisp and alive, and the resulting read-and-edit story.

\subsection{Training for interpretability, and objective-induced collapse}
\label{sec:bg-training}
Inducing interpretable structure during training, rather than post hoc, has precedent: Concept Bottleneck Models align an intermediate layer to labeled concepts \citep{koh2020concept}, Self-Explaining Neural Networks regularize explanations for stability \citep{alvarez2018senn}, and end-to-end sparse dictionary learning folds a task term into an auxiliary dictionary \citep{braun2024e2esae}. Unlike concept bottlenecks we require no concept labels, and unlike dictionary methods the structure lives in the primary computation rather than a separate decoder. Our crispness pressure is related to soft binarization and straight-through quantization \citep{courbariaux2016binarized,bengio2013ste,yin2019ste}, but targets legible Boolean operators rather than compression.

The failure we diagnose and repair is a collapse of variance, connecting to the anti-collapse literature in self-supervised learning. VICReg prevents representational collapse with a per-dimension variance hinge $\relu(\gamma-\mathrm{std})$ paired with covariance decorrelation \citep{bardes2022vicreg}; related mechanisms include redundancy reduction \citep{zbontar2021barlow}, feature whitening \citep{ermolov2021whitening}, and the predictor/stop-gradient schemes that motivated explicit variance terms \citep{grill2020byol,chen2021simsiam}. Our variance floor shares VICReg's hinge shape. The difference is what it repairs: VICReg substitutes for contrastive negatives against an invariance loss, whereas our floor counteracts a collapse \emph{induced by our own crispness term}, which the identity $\mathrm{mean}\,v(1-v)=\mu(1-\mu)-\mathrm{var}$ reveals to be a variance-minimizer that cannot distinguish a live contextual operator from a dead constant. The resulting dead unit is the crispness-induced analogue of the classical dying-ReLU state \citep{lu2020dying}. Optimizing a crispness proxy that destroys the target it stands for is a specification-gaming pattern \citep{manheim2018goodhart,jacovi2020faithful}, and the identity makes precise why a variance floor is the minimal correction.

\subsection{Reading hidden states in vocabulary space}
\label{sec:bg-vocab}
Projecting intermediate states through the unembedding --- the logit lens \citep{nostalgebraist2020logitlens} --- is a standard readout, with a well-documented failure at middle layers. The tuned lens attributes this to representations living in a rotated basis and recovers predictions with a learned per-layer affine map, framed as iterative inference \citep{belrose2023tunedlens}; the iterative-refinement view of residual networks is older still \citep{jastrzebski2018iterative,greff2017unrolled}. Feed-forward writes decode to vocabulary as key-value memories that promote concepts \citep{geva2021kvmemories,geva2022promoting}, and per-component logit attribution follows from the residual stream being a linear channel with no privileged basis \citep{elhage2021framework}, a basis that is nonetheless weakly privileged in practice \citep{elhage2023privilegedbasis}. Learned lenses have been applied at the granularity of individual attention heads \citep{sakarvadia2023attentionlens}, whole parameter matrices \citep{dar2023embeddingspace}, and future tokens \citep{pal2023futurelens}. We use these established lenses; our reading results are their application to legible-by-construction units and the detection-versus-naming distinction of Section~\ref{sec:understand}, not a new readout.

\subsection{Superposition, polysemanticity, and non-lexical features}
\label{sec:bg-superposition}
Superposition explains why a crisp unit can fire on an incoherent set of tokens: networks pack more features than neurons as near-orthogonal directions, and polysemanticity is the direct consequence \citep{elhage2022toy}. The polytope lens reframes this geometrically, arguing that meaning is monosemantic over activation \emph{regions} even where directions are polysemantic \citep{black2022polytope} --- the local-coherence view our per-input reads inherit. Sparse autoencoders recover largely monosemantic features from polysemantic neurons \citep{bricken2023monosemanticity,cunningham2023sparse}, at scale surfacing abstract, multilingual, non-lexical concepts \citep{templeton2024scaling,gao2024scaling,lieberum2024gemmascope}. That concepts occupy directions not aligned to any token is predicted by the linear representation hypothesis, which separates the context and unembedding geometries \citep{park2024linear}, and is documented directly for function vectors that steer yet do not decode \citep{nadaf2026steerable} and for the unexplained residual of dictionary methods \citep{engels2024darkmatter,engels2024multidim}. Whether a unit is nameable is operationalized by automated interpretability \citep{bills2023neurons,mu2020compositional,kopf2025prism,eleuther2024autointerp}. Against this backdrop our contribution is specific: because our units are crisp by construction, their per-input meaning is recoverable cheaply, and we report the layerwise trajectory by which a concept that is non-lexical in mid-stack becomes token-decodable only in the final layers.

\subsection{Model editing and activation steering}
\label{sec:bg-editing}
The mechanism we use to edit is standard. Feed-forward layers are editable key-value memories \citep{geva2021kvmemories,geva2022promoting}; knowledge is localizable to individual neurons \citep{dai2022knowledge} and to mid-layer feed-forward sites whose \emph{layer-native} value vector ROME and MEMIT solve for --- not the token embedding \citep{meng2022rome,meng2023memit}. Activation steering adds a direction obtained as a difference of means over contrast contexts \citep{turner2023actadd,rimsky2024caa,zou2023repe}, sufficient to move behavior with little collateral \citep{arditi2024refusal}; behaviors also compose as task and function vectors \citep{ilharco2023task,todd2024function,hendel2023taskvectors}, and causal mediation localizes them \citep{vig2020causal,wang2023ioi,zhang2024patching}. Our deep edits are difference-of-means writes in this native frame. Two results bound the contribution: interpretability is only a weak proxy for steering utility \citep{wu2025interpretability}, and more monosemantic units incur less collateral \citep{templeton2024scaling,neuronlens2025ranges}. We differ in that our units are made crisp and sparse \emph{by construction} rather than discovered, are edited in-weights rather than clamped at inference, and support a single-unit conjunctive $A\cap B$ edit that a single-preactivation neuron cannot represent.

%% file: architecture.tex
\section{The Legible Substrate}
\label{sec:architecture}

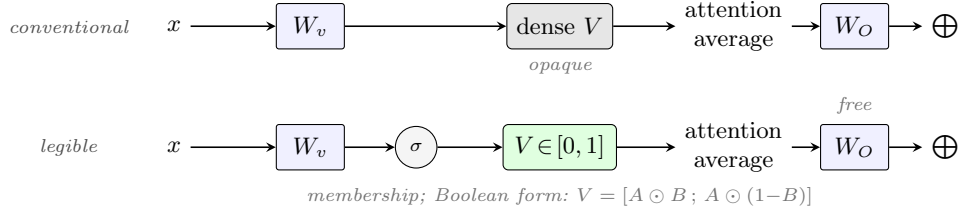
\begin{figure}[htbp]
\centering
\input{fig_arch.tikz}
\caption{The legible substrate, on both sublayers. \emph{Feed-forward:} a conventional layer builds a dense hidden vector with no per-unit meaning; the legible layer instead forms two sigmoid-bounded operands $A,B\in[0,1]$ and combines them by fuzzy set operations --- intersection $A\cap B$ (``$A$ and $B$'') and set-difference $A\setminus B$ (``$A$ and-not $B$'') --- then writes with a free $W_o$. \emph{Attention value:} conventionally a dense, unconstrained vector; legibly a bounded membership $V\in[0,1]$ (or the same Boolean set-operator form), averaged by the attention map and written with a free $W_O$. In both sublayers what a unit \emph{detects} is bounded and named, while what it \emph{writes} is left free.}
\label{fig:arch}
\end{figure}

The models we train are built from components that are legible \emph{by construction}: their operators are bounded and named, so a coordinate carries a fixed meaning a human can state, rather than a dense activation whose meaning must be reconstructed after the fact. The feed-forward units are those of \citet{oskin2026boolglu} and the attention value heads are those of \citet{oskin2026attention}; we recap both here so the paper is self-contained, and refer to those works for the fuller development. Both share one principle: \emph{constrain what a unit detects, and leave what it writes free} (Figure~\ref{fig:arch}). This section states the operators, the pressures that shape them, and the metric by which we call a unit legible; Section~\ref{sec:train} is about the objective that makes the whole thing hold together.

\subsection{Legible feed-forward units}
\label{sec:arch-ffn}
A conventional feed-forward layer computes $\mathrm{FFN}(x)=W_o\,\GELU(W_{\mathrm{in}}x)$, whose hidden units have no reason to be individually meaningful. The legible layer replaces the hidden activation with fuzzy set operations on two banks of sigmoid-bounded operands. From the input $x$ it forms
\begin{equation}
A=\sig(W_a x)\in[0,1]^{m},\qquad B=\sig(W_b x)\in[0,1]^{m},
\end{equation}
and constructs the hidden vector as the concatenation of a fuzzy intersection and a negation-capable set-difference,
\begin{equation}
h \;=\; \big[\,\underbrace{A\odot B}_{\text{$A$ \textsc{and} $B$}}\;;\;\underbrace{A\odot(1-B)}_{\text{$A$ \textsc{and-not} $B$}}\,\big],
\qquad \mathrm{FFN}(x)=W_o\, h .
\label{eq:ffn}
\end{equation}
A first-block coordinate reads as ``both $A$ and $B$ present,'' a second-block coordinate as ``$A$ present, $B$ absent''; the blocks are asymmetric since $A\setminus B\neq B\setminus A$. Because $W_a,W_b$ together match the size of the projection they replace, the layer is parameter-neutral. For trainability a fraction $\rho$ of hidden units are kept as conventional $\GELU$ and only the remainder are set operators \citep{oskin2026boolglu}; we write the named fraction as $1-\rho$. A set-operator unit can become a $\GELU$ --- pass its first operand's pre-activation $W_a x$ through the nonlinearity --- but only at a $50\%$ capacity cost: the second projection $W_b$ then idles, so the resulting $\GELU$ carries twice the input-projection weight of a native $\GELU$ neuron. The reserved fraction $\rho$ is thus a parameter tax as much as a dilution of the legible substrate, a cost Section~\ref{sec:train-lcf} removes by learning the fraction per unit. The write $W_o$ is left unconstrained: a unit is read on the \emph{detection} side --- which operands fired --- not by decoding what it writes.

\subsection{Legible attention value heads}
\label{sec:arch-attn}
The same principle applies to the attention value, the opaque half of a head \citep{elhage2021framework}: the attention map already shows \emph{where} a head reads, but the value it moves is a dense unconstrained vector. We constrain that value before the output projection $W_O$ and leave $W_O$ free. In the \emph{membership} form a head's value is simply bounded,
\begin{equation}
V_h=\sig(X W_v^h)\in[0,1]^{T\times d_h},\qquad O_h=A_h V_h,
\label{eq:membership}
\end{equation}
so each coordinate is a fuzzy membership --- the degree to which a named feature is present --- and attention forms a fuzzy-weighted average of memberships over the attended context. In the \emph{Boolean} form the value projection is split into two half-width banks $A=\sig(XW_a^h)$, $B=\sig(XW_b^h)$ and the value is the same set-operator construction as the FFN,
\begin{equation}
V_h=\big[\,A\odot B\;;\;A\odot(1-B)\,\big]\in[0,1]^{T\times d_h}.
\label{eq:boolean}
\end{equation}
Both forms are parameter-neutral. A layer may mix bounded heads with conventional ones; we use the fully-bounded ($100\%$) setting throughout, coupling Boolean attention (Eq.~\ref{eq:boolean}) to the legible FFN (Eq.~\ref{eq:ffn}) to make an end-to-end model in which both what a layer \emph{moves} and what it \emph{combines} are named operations. Constraining the value rather than the write is essential: pinning the post-$W_O$ write to the vocabulary instead hands the head a gradient path to the unigram prior and collapses it to a context-independent bias \citep{oskin2026attention}.

\subsection{Selectivity pressures}
\label{sec:arch-pressure}
A bounded operator is not automatically legible. A readable detector should fire on a \emph{sparse, specific} set of contexts, and it should fire \emph{decisively} --- near $0$ or near $1$ rather than in the ambiguous middle. These are two properties, imposed by two small penalties on the bounded operands (the atoms we want to name --- $A$ and $B$ --- not the combined output), added to the language-model loss. A \emph{sparsity} pressure
\begin{equation}
\mathcal{L}_{\mathrm{s}}=\lambda_{\mathrm{s}}\,\textstyle\sum_{t,c}|v_{t,c}|
\end{equation}
pushes each value toward $0$; it has teeth only because the value is bounded (with a free $W_O$, a penalty on an unbounded value is rescaled away). A \emph{crispness} pressure
\begin{equation}
\mathcal{L}_{\mathrm{c}}=\lambda_{\mathrm{c}}\,\textstyle\sum_{t,c} v_{t,c}(1-v_{t,c})
\label{eq:crisp}
\end{equation}
is maximal at $\tfrac12$ and zero at the rails, driving each value decisively toward the nearer rail without preferring off. We use $\lambda_{\mathrm{s}}=10^{-3}$, $\lambda_{\mathrm{c}}=3\times10^{-3}$. Which combination is right is not universal --- for a lone membership value sparsity is what carves out a selective detector, whereas a conjunction supplies its own sparsity and wants crispness alone \citep{oskin2026attention}. The failure mode of the crispness pressure, and its repair, is the subject of Section~\ref{sec:train}.

\subsection{When is a unit legible?}
\label{sec:arch-metric}
\begin{figure}[htbp]
\centering
\includegraphics[width=\textwidth]{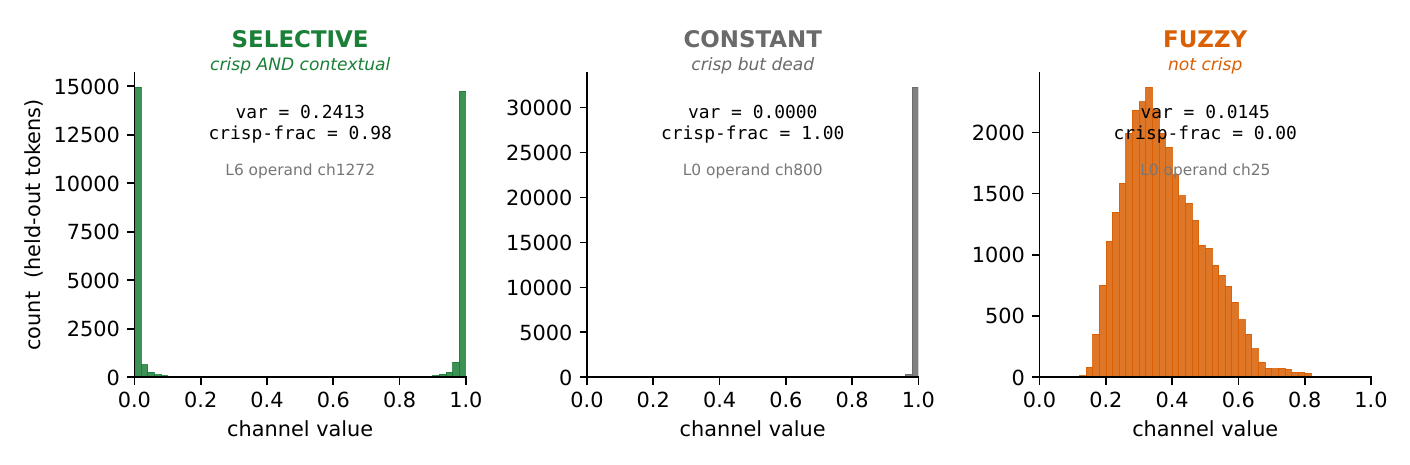}
\caption{The three metric states, each a real feed-forward operand channel of the annealed model over held-out tokens. \SELECTIVE{} (crisp \emph{and} contextual): mass at \emph{both} rails (variance $0.24$). \CONSTANT{} (crisp but dead): a single spike at one rail (variance ${\sim}0$). \textsc{fuzzy} (not crisp): mass in the ambiguous middle. Only \SELECTIVE{} is a decisive, input-dependent detector.}
\label{fig:metric}
\end{figure}

We call a bounded operand channel \emph{legible} when it is both crisp and contextual, and we measure this on held-out text. Over a sample of contexts let a channel have per-token values with mean $\mu_c$ and variance $\mathrm{var}_c$. The channel is \emph{crisp} when its values concentrate at the rails (the fraction within $0.1$ of $\{0,1\}$ exceeds one half), and \emph{contextual} when it varies across inputs, $\mathrm{var}_c\ge\tau$ with $\tau=0.003$. This gives three states (Figure~\ref{fig:metric}):
\begin{itemize}
\item \SELECTIVE{} --- crisp \emph{and} contextual: a decisive, input-dependent detector, the readable case;
\item \CONSTANT{} --- crisp but $\mathrm{var}_c<\tau$: decisive but input-\emph{independent}, a dead constant;
\item \textsc{fuzzy} --- not crisp: a soft gate that never commits.
\end{itemize}

\SELECTIVE{} is the target state. It is a property of the \emph{detection} side and is independent of any choice of readout basis. The distinction between \SELECTIVE{} and \CONSTANT{} --- both perfectly crisp, separated only by whether the unit still responds to its input --- is exactly what a crispness pressure alone cannot see, as Section~\ref{sec:train} shows.

One caveat travels with the count wherever we report it. The pressures of Section~\ref{sec:arch-pressure} settle the crisp population \emph{just above} the variance floor --- a typical channel's across-context variance is only a small multiple of $\tau$ --- so the \SELECTIVE{} \emph{count} is a hard threshold applied to mass sitting right at the cutoff. A small, transient drift in variance can carry many channels across the line, and the count can swing by tens of points between training checkpoints while the crispness underneath barely moves. We therefore report \SELECTIVE{} fractions only at convergence (end of epoch), and pair the count with the mean \emph{crisp fraction} --- the average share of a channel's values within $0.1$ of a rail --- which carries the same crispness information without the threshold fragility and is stable across checkpoints (${\approx}0.90$ on the annealed model). The crisp fraction is the continuous form of the crisp criterion; the \SELECTIVE{} count adds only the variance test that a single mid-training snapshot can misread. Section~\ref{sec:dial} takes this up directly.

\subsection{Setup}
\label{sec:arch-setup}
All models are $125$M-parameter, $12$-layer, $768$-wide transformers with learned positional embeddings, trained for one epoch on the same open-web corpus and tokenizer, with identical optimizer and schedule; the conventional \textsc{gelu} model is architecturally identical except for the value transform and FFN. We report language-model quality with dev perplexity, LAMBADA \citep{paperno2016lambada}, and BLiMP \citep{warstadt2020blimp} via the standard harness \citep{gao2023lmeval}, and legibility with the \SELECTIVE{}/\CONSTANT{} statistics above. All results are single seed at this scale.

%% file: fig_arch.tikz
% F-arch — the legible substrate. Two groups (feed-forward, attention value), each conventional vs legible.
% Self-contained (basic tikz, default arrows, explicit coords; \GELU/\sig from the paper preamble).
\begin{tikzpicture}[
  font=\small,
  proj/.style={draw, rounded corners=1pt, minimum height=6mm, minimum width=9mm, fill=blue!6},
  act/.style={draw, circle, inner sep=0.4pt, minimum size=5.5mm, fill=black!4},
  bval/.style={draw, rounded corners=2pt, minimum height=6mm, minimum width=15mm, fill=green!12},
  opaque/.style={draw, rounded corners=2pt, minimum height=6mm, minimum width=14mm, fill=black!10},
  comb/.style={draw, rounded corners=2pt, minimum height=10mm, minimum width=19mm, fill=green!12, align=center},
  note/.style={font=\scriptsize\itshape, text=black!55},
  grp/.style={font=\small\bfseries},
  lane/.style={font=\scriptsize\itshape, text=black!70},
  ar/.style={->, >=stealth, semithick}]

% ============ FEED-FORWARD ============
\node[grp] at (-1.5,4.0) {feed-forward};
% conventional FFN
\node[lane] at (-1.5,3.2) {conventional};
\node (cx)   at (-0.1,3.2) {$x$};
\node[proj]   (cwin) at (1.7,3.2) {$W_{\mathrm{in}}$};
\node[act]    (cg)   at (3.4,3.2) {\scriptsize$\GELU$};
\node[opaque, minimum width=17mm] (ch) at (6.0,3.2) {dense hidden};
\node[proj]   (cwo)  at (8.9,3.2) {$W_o$};
\node         (cadd) at (10.1,3.2) {$\bigoplus$};
\draw[ar](cx)--(cwin); \draw[ar](cwin)--(cg); \draw[ar](cg)--(ch); \draw[ar](ch)--(cwo); \draw[ar](cwo)--(cadd);
\node[note] at (6.0,2.65) {no per-unit meaning};
% legible FFN
\node[lane] at (-1.5,1.0) {legible};
\node (lx) at (-0.1,1.0) {$x$};
\node[proj] (lwa) at (1.7,1.75) {$W_a$};
\node[proj] (lwb) at (1.7,0.25) {$W_b$};
\node[act]  (lsa) at (3.1,1.75) {\scriptsize$\sig$};
\node[act]  (lsb) at (3.1,0.25) {\scriptsize$\sig$};
\node[bval] (la)  at (4.9,1.75) {$A\!\in\![0,1]$};
\node[bval] (lb)  at (4.9,0.25) {$B\!\in\![0,1]$};
\node[comb] (lc)  at (7.0,1.0) {$A\odot B$\\[2pt]$A\odot(1{-}B)$};
\node[proj] (lwo) at (8.9,1.0) {$W_o$};
\node       (ladd) at (10.1,1.0) {$\bigoplus$};
\draw[ar](lx)--(lwa); \draw[ar](lx)--(lwb);
\draw[ar](lwa)--(lsa); \draw[ar](lwb)--(lsb);
\draw[ar](lsa)--(la); \draw[ar](lsb)--(lb);
\draw[ar](la)--(lc); \draw[ar](lb)--(lc);
\draw[ar](lc)--(lwo); \draw[ar](lwo)--(ladd);
\node[note, align=center] at (7.0,-0.3) {``$A$ and $B$'' \ ; \ ``$A$ and-not $B$''};
\node[note] at (8.9,1.55) {free};

% ============ ATTENTION VALUE ============
\draw[black!15] (-2.4,-0.75) -- (10.9,-0.75);
\node[grp] at (-1.5,-1.2) {attention value};
% conventional attention value
\node[lane] at (-1.5,-2.0) {conventional};
\node (cax) at (-0.1,-2.0) {$x$};
\node[proj] (cawv) at (1.7,-2.0) {$W_v$};
\node[opaque] (cav) at (5.0,-2.0) {dense $V$};
\node[align=center] (caat) at (7.3,-2.0) {attention\\average};
\node[proj] (cawo) at (8.9,-2.0) {$W_O$};
\node (caadd) at (10.1,-2.0) {$\bigoplus$};
\draw[ar](cax)--(cawv); \draw[ar](cawv)--(cav); \draw[ar](cav)--(caat); \draw[ar](caat)--(cawo); \draw[ar](cawo)--(caadd);
\node[note] at (5.0,-2.55) {opaque};
% legible attention value
\node[lane] at (-1.5,-3.6) {legible};
\node (ax) at (-0.1,-3.6) {$x$};
\node[proj] (awv) at (1.7,-3.6) {$W_v$};
\node[act]  (asv) at (3.1,-3.6) {\scriptsize$\sig$};
\node[bval] (av)  at (5.0,-3.6) {$V\!\in\![0,1]$};
\node[align=center] (aat) at (7.3,-3.6) {attention\\average};
\node[proj] (awo) at (8.9,-3.6) {$W_O$};
\node       (aadd) at (10.1,-3.6) {$\bigoplus$};
\draw[ar](ax)--(awv); \draw[ar](awv)--(asv); \draw[ar](asv)--(av); \draw[ar](av)--(aat); \draw[ar](aat)--(awo); \draw[ar](awo)--(aadd);
\node[note, align=center] at (5.0,-4.25) {membership; Boolean form:\ $V=[A\odot B\,;\,A\odot(1{-}B)]$};
\node[note] at (8.9,-3.05) {free};
\end{tikzpicture}

%% file: train.tex
\section{Training Legible Models}
\label{sec:train}

The crispness pressure of Eq.~\ref{eq:crisp} is what turns a bounded operand into a decisive, readable detector. It also has a failure mode. Past a point it collapses the operators it was meant to sharpen into \emph{dead constants}: units that are perfectly crisp and completely uninformative. We show why this happens with a two-line identity, repair it with a variance floor that turns the \SELECTIVE{} metric into a training objective, and give a learned-fraction variant that lets a unit revert to $\GELU$ where logic is not needed. The repair recovers both legibility and quality.

\subsection{The collapse}
\label{sec:train-collapse}
Apply the crispness pressure to the feed-forward operands of the fully-Boolean model and, past a point, the operands go crisp but dead. On held-out text only $2\%$ of feed-forward operands are \SELECTIVE{} while most are \CONSTANT{} --- crisp, but pinned to a rail regardless of input --- and quality craters: LAMBADA perplexity rises to $240$ against $153$ for the same model with the feed-forward operands left unpressured. The damage does not stay local. Because the collapsed units still write into the shared residual stream, the attention operands degrade with them, their \SELECTIVE{} fraction falling from $27\%$ to $20\%$. A pressure meant to sharpen the operators instead switches them off.

\subsection{The problem}
\label{sec:train-identity}
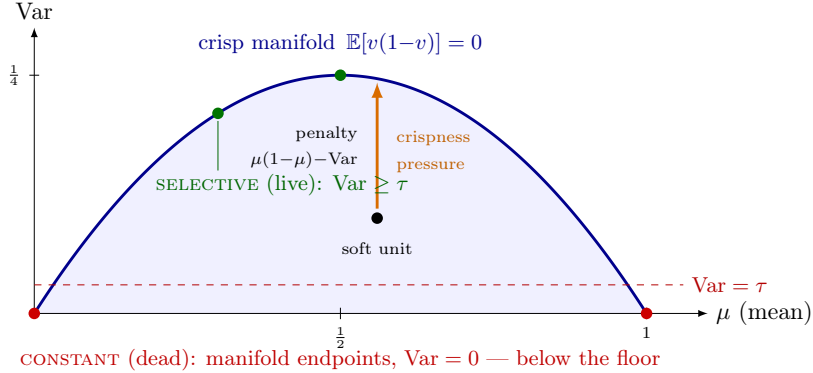
\begin{figure}[htbp]
\centering
\scalebox{0.9}{\input{fig_manifold.tikz}}
\caption{Why the crispness pressure cannot sharpen an operator without risking its death. The penalty $\mathbb{E}[v(1-v)]=\mu(1-\mu)-\mathrm{Var}$ is the vertical gap below the parabola $\mathrm{Var}=\mu(1-\mu)$; driving it to zero places a unit anywhere on that \emph{crisp manifold}, where a live \SELECTIVE{} detector ($\mathrm{Var}\ge\tau$) and a dead \CONSTANT{} unit (the endpoints, $\mathrm{Var}=0$) are indistinguishable to it. The variance floor (dashed) prices out the endpoints, leaving the live interior as the only zero-loss state. The variance axis is not to scale: the true $\tau=0.003$ lies just above the axis.}
\label{fig:manifold}
\end{figure}

For a bounded value $v\in[0,1]$ with mean $\mu$ and variance $\mathrm{var}$ over a batch, the crispness penalty of Eq.~\ref{eq:crisp} decomposes exactly:
\begin{equation}
\mathbb{E}\big[v(1-v)\big] \;=\; \mu(1-\mu)-\mathrm{var}.
\label{eq:identity}
\end{equation}
Since $\mathrm{var}\le\mu(1-\mu)$ the quantity is nonnegative, and it reaches zero on the \emph{entire} crisp manifold (Figure~\ref{fig:manifold}): every operator whose mass sits at the rails drives it to zero, whether it is a live contextual detector ($\mu$ interior, $\mathrm{var}>0$, mass split between the rails across inputs) or a dead constant ($\mu$ at a rail, $\mathrm{var}=0$). The penalty cannot tell the two apart. Worse, the readout $W_o$ is free, so crispness carries no obligation to be useful; it is minimized as a pure tax, and the cheapest way to pay it is a constant. Once there, the saturated sigmoid gives almost no gradient, and the dead state is absorbing. The crispness pressure is, by Eq.~\ref{eq:identity}, a variance-minimizer in disguise --- and minimizing the variance of a detector is precisely how you kill it.

\subsection{The fix: a variance floor}
\label{sec:train-ctx}
\begin{figure}[htbp]
\centering
\includegraphics[width=0.72\textwidth]{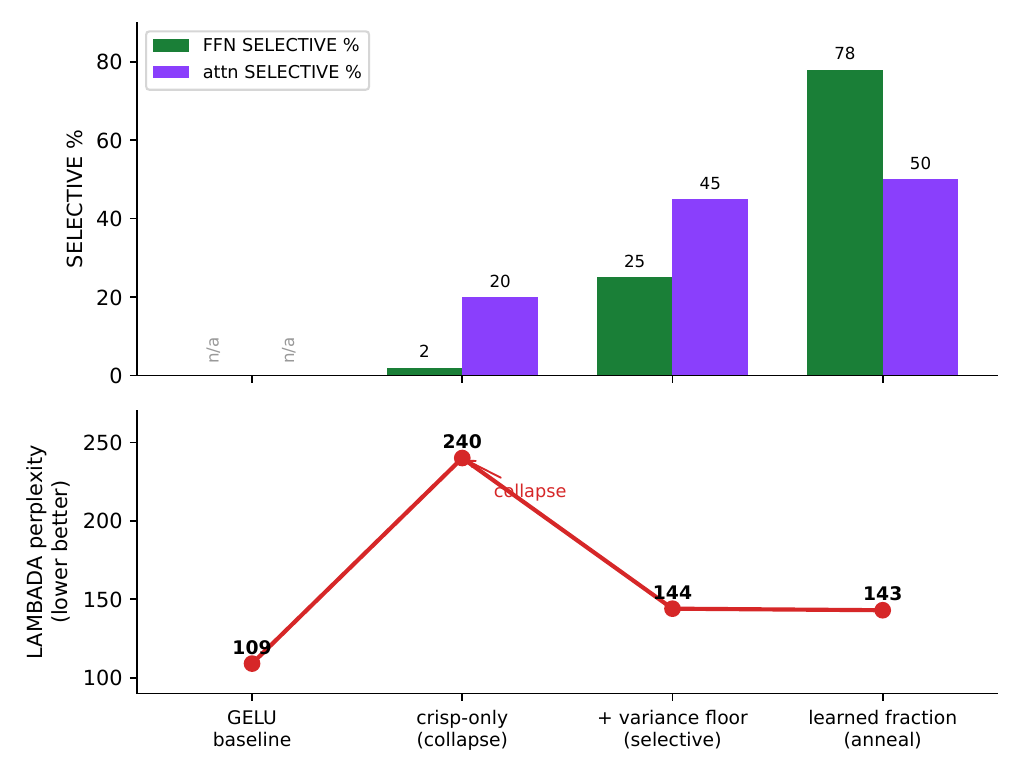}
\caption{Repairing the crispness collapse (fully-Boolean $12$-layer model). \emph{Top:} the fraction of \SELECTIVE{} feed-forward and attention operands across the four configurations. \emph{Bottom:} LAMBADA perplexity. Crispness pressure alone collapses the operators (feed-forward \SELECTIVE{} to $2\%$, perplexity to $240$); the variance floor (\emph{selective}) and the learned crisp fraction (\emph{anneal}) recover quality, and the learned fraction reaches the highest legibility ($78\%$). The \textsc{gelu} baseline has no bounded operands (n/a).}
\label{fig:collapsefix}
\end{figure}

The identity also says how to fix it: forbid the zero-variance solution. We add a per-channel hinge that is positive when a channel's variance is below the \SELECTIVE{} threshold and zero once it clears it,
\begin{equation}
\mathcal{L}_{\mathrm{ctx}} \;=\; \lambda_{\mathrm{ctx}}\;\mathrm{mean}_c\;\relu\!\Big(1-\frac{\mathrm{var}_c}{\tau}\Big),\qquad \tau=0.003.
\label{eq:ctx}
\end{equation}
This is the \SELECTIVE{} metric of Section~\ref{sec:arch-metric} turned into a loss: paired with the crispness term, the unique zero-loss state is crisp \emph{and} contextual, and a dead constant now pays $\lambda_{\mathrm{ctx}}$. Three properties matter. It adds \emph{no parameters}. It floors \emph{variance}, not mean or firing rate, which is what protects a rare-but-real detector --- a channel active on $1\%$ of tokens has $\mathrm{var}\approx0.0099>\tau$ and is left alone, whereas a mean-based floor would force it on. And because a saturated dead unit revives only slowly, the floor must be applied from the first step, where it prevents deaths rather than reversing them. With $\lambda_{\mathrm{ctx}}=3\times10^{-3}$ on both sublayers --- the model we call \emph{selective} --- the collapse reverses on every axis (Figure~\ref{fig:collapsefix}): feed-forward \SELECTIVE{} rises $2\%\!\to\!25\%$ and \CONSTANT{} falls sharply; attention \SELECTIVE{} rises $20\%\!\to\!45\%$; and quality recovers with legibility --- dev perplexity $23.2\!\to\!20.1$, LAMBADA perplexity $240\!\to\!144$, BLiMP $0.782\!\to\!0.819$.

The floor shares its hinge shape with the variance term of VICReg \citep{bardes2022vicreg}. What is specific here is the problem it solves: not the informational collapse of self-supervised learning, but a collapse \emph{induced by another term in our own objective}, which the identity of Eq.~\ref{eq:identity} identifies as a variance-minimizer, and which the floor counteracts to select the crisp-and-contextual corner that neither term reaches alone.

\subsection{Where the floor is needed: a sublayer dissociation}
\label{sec:train-ablation}
\begin{figure}[htbp]
\centering
\includegraphics[width=0.85\textwidth]{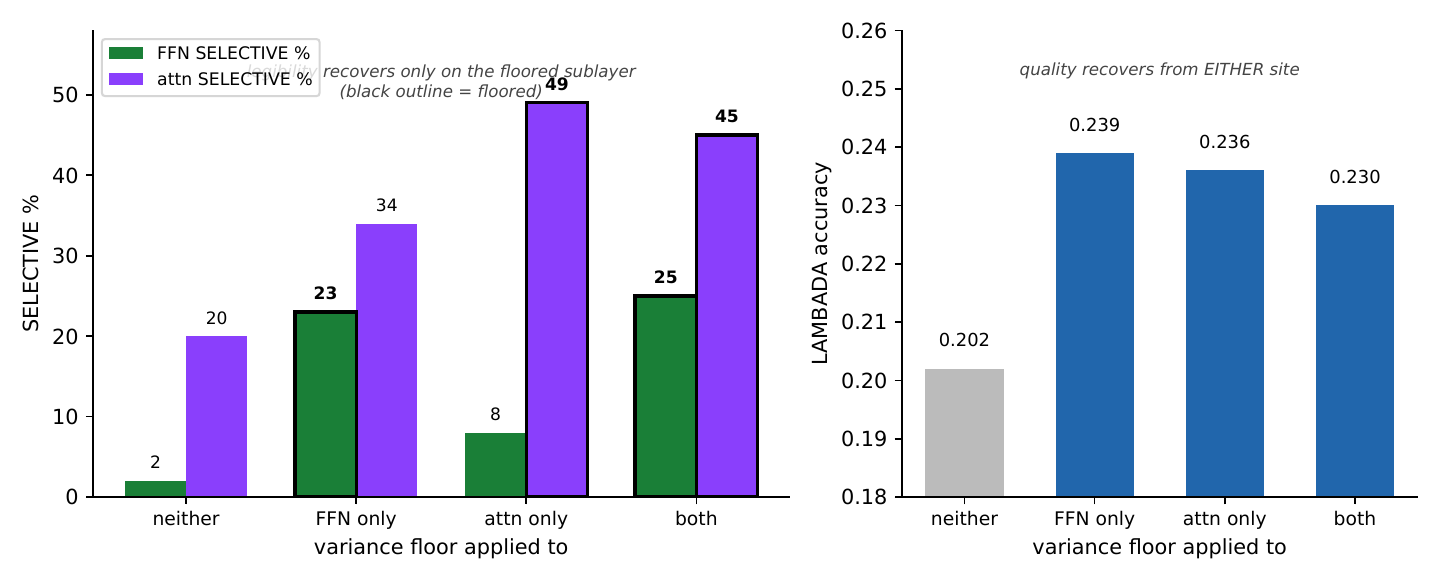}
\caption{Variance floor on each sublayer ($12$L). \emph{Left:} \SELECTIVE{} fractions --- each sublayer's legibility recovers only when its \emph{own} floor is applied (black outline marks the floored sublayer): a \emph{local} effect. \emph{Right:} LAMBADA accuracy recovers from \emph{either} floor: a \emph{global} effect. Quality is a coupled property of the residual stream; legibility is local to each sublayer's operators.}
\label{fig:ablation}
\end{figure}

Placing the floor on one sublayer at a time separates its two effects (Figure~\ref{fig:ablation}). \emph{Quality recovery is global}: flooring \emph{either} sublayer recovers most of the collapse (LAMBADA accuracy $0.20\!\to\!0.24$, dev perplexity $23.2\!\to\!20.4$), consistent with the collapse spreading through the shared residual --- healing one end helps the whole stream --- though BLiMP climbs only when both are floored. \emph{Legibility is local}: each floor drives \SELECTIVE{} chiefly on its own sublayer (the feed-forward floor lifts feed-forward \SELECTIVE{} $2\%\!\to\!23\%$ but attention only to $34\%$; the attention floor lifts attention $20\%\!\to\!49\%$ but feed-forward only to $8\%$). The \SELECTIVE{} count fluctuates by a few points near the variance threshold (Section~\ref{sec:arch-metric}), so what the dissociation rests on is each floor's lift on its \emph{own} sublayer, not the small wobble in the attention count across configurations (here $49\%$, and $45\%$ with both floored). Only flooring both is legible on both sublayers while reaching the top BLiMP. Quality is a coupled property of the residual stream; legibility is a local property of each sublayer's operators.

\subsection{Learning the operator fraction instead of reserving it}
\label{sec:train-lcf}
The legible feed-forward layer carries a hand-set hyperparameter: a fraction of its hidden units is held back as conventional $\GELU$ for trainability \citep{oskin2026boolglu}, a fraction as large as one half. We replace that fixed reservation with a per-unit \emph{learned} choice. Each unit interpolates, by a gate $\alpha=\sig(\theta)$, between its set operator and a $\GELU$ computed from the same pre-activation, so the choice costs one scalar per unit and nothing else: $\alpha\!\to\!1$ is a crisp operator, $\alpha\!\to\!0$ a reverted nonlinearity. A crisp-first initialization $\alpha\approx0.9$ starts every unit in the operator basin while leaving a small $\GELU$ path open for trainability, and an output-norm-weighted usage tax $\lambda(1-\alpha)\lVert W_o[:,u]\rVert$ prices reverting, so a dead operator cannot escape the tax by writing nothing.

\begin{figure}[htbp]
\centering
\includegraphics[width=0.75\textwidth]{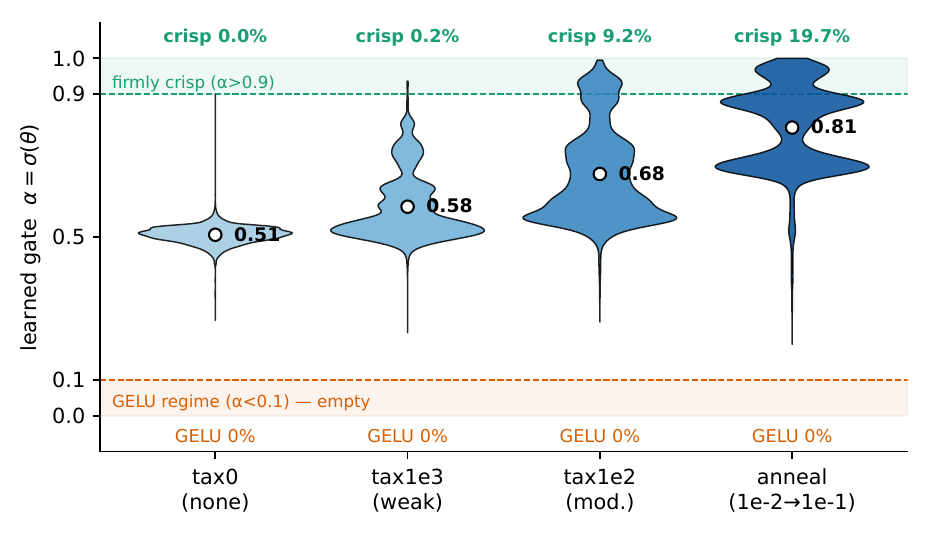}
\caption{Distribution of the learned gate $\alpha=\sig(\theta)$ across the four runs. The pure-\GELU{} band ($\alpha<0.1$, bottom) is empty in every run, including the untaxed null; the crisp band ($\alpha>0.9$, top) fills as the tax rises. The model is never handed a reserved \GELU{} partition and never forms one on its own.}
\label{fig:alpha}
\end{figure}

Given the choice, the model reserves no unit for pure $\GELU$. The fully-reverted bucket ($\alpha<0.1$) is empty in every run --- including the null run with the tax switched off --- and across all runs the smallest gate anywhere is $\alpha\approx0.2$ (Table~\ref{tab:lcf}). The reversion to an all-$\GELU$ layer never occurs, and the hand-set partition is unnecessary: the crisp-first initialization holds the operator basin without a budget floor, and the per-unit gate supplies the trainability cushion the fixed fraction used to, distributed as a thin per-unit blend rather than a reserved block. The tax then sets how far the population crispens. Firmly-crisp units ($\alpha>0.9$) grow from $0\%$ to $20\%$ under an annealed schedule ($10^{-2}\!\to\!10^{-1}$) and the mean gate rises to $0.81$ (Figure~\ref{fig:alpha}); weighted by output norm --- the computation that actually drives predictions --- crisp operators carry $0.87$ of the load. The tax is not only a legibility knob: its lightest setting already improves quality over the untaxed null (Table~\ref{tab:lcf}). The annealed model, our showcase, reaches the highest legibility of any configuration --- $78\%$ of feed-forward operands \SELECTIVE{} at only $11\%$ \CONSTANT{}, and $50\%$ of attention operands \SELECTIVE{} --- at quality on par with the variance-floor model (Figure~\ref{fig:collapsefix}). The reservation the architecture required is gone; what remains is a small learned residual, and it does not fall to zero --- the layer settles as a blend rather than a fully-Boolean one, consistent with the trainability ceiling of Appendix~\ref{sec:appendix-trainability}.

\begin{table}[htbp]
\centering
\begin{tabular}{lcccccc}
\toprule
$\GELU$ tax & mean $\alpha$ & min $\alpha$ & $\alpha<0.1$ & $\alpha>0.9$ & wtd.\ crisp & LAMBADA / BLiMP \\
\midrule
$0$ (null) & 0.51 & 0.27 & 0\% & 0.0\%  & 0.51 & 0.223 / 0.795 \\
$10^{-3}$  & 0.59 & 0.23 & 0\% & 0.2\%  & 0.53 & 0.249 / 0.810 \\
$10^{-2}$  & 0.68 & 0.26 & 0\% & 9.2\%  & 0.61 & 0.237 / 0.808 \\
anneal     & 0.81 & 0.20 & 0\% & 19.7\% & 0.87 & 0.232 / 0.809 \\
\bottomrule
\end{tabular}
\caption{Learning the operator fraction ($12$L). Given a per-unit learned gate $\alpha$, the model reserves \emph{no} unit for pure $\GELU$: the $\alpha<0.1$ bucket is empty in every run, including the untaxed null, and the smallest gate anywhere is $\approx0.2$. The tax sets how far the population crispens; ``wtd.\ crisp'' is the output-norm-weighted crisp fraction --- the share of load-bearing computation carried by crisp operators.}
\label{tab:lcf}
\end{table}

\subsection{Pressures that do not work}
\label{sec:train-failures}
Several natural alternatives fail, and the failures locate the mechanism. \emph{Crispness alone} is the collapse of Section~\ref{sec:train-collapse}. \emph{A single-sided floor} recovers quality but leaves the unfloored sublayer's operators dead (Section~\ref{sec:train-ablation}), so a model that looks healthy on perplexity can be half-collapsed on legibility. \emph{Too strong a floor} ($\lambda_{\mathrm{ctx}}=10^{-2}$) over-pushes variance and is slightly worse than $3\times10^{-3}$ on quality. \emph{No pressure at all} on the learned fraction (the null tax) leaves units at a soft $\alpha\approx0.5$ with no crisp population --- stable and adequate on quality, but producing nothing to read. And the named fraction is not free without limit: past roughly half-Boolean the feed-forward layer meets a late-training instability \citep{oskin2026boolglu}, so legibility is free at parity only \emph{up to} a partition, not across a fully Boolean layer. The variance floor and the learned crisp fraction are the two pressures that hold a live, crisp, contextual population without either collapse.

%% file: fig_manifold.tikz
% F2 — the Bernoulli / crisp manifold: why crispness pressure cannot tell dead from alive.
% Data units: x = mu in [0,1], y = Var in [0,0.30].  Self-contained (basic tikz + amsmath).
\begin{tikzpicture}[xscale=9,yscale=14,>=latex,
    dot/.style={circle,fill,inner sep=1.7pt}]
  % feasible region  Var <= mu(1-mu)
  \fill[blue!6,domain=0:1,samples=120] plot (\x,{\x*(1-\x)}) -- (1,0) -- (0,0) -- cycle;
  % axes
  \draw[->] (0,0) -- (1.10,0) node[right] {$\mu$ (mean)};
  \draw[->] (0,0) -- (0,0.30) node[above] {$\mathrm{Var}$};
  \draw (0.5,0.006) -- (0.5,-0.006) node[below] {\scriptsize $\tfrac12$};
  \draw (1,0.006)   -- (1,-0.006)   node[below] {\scriptsize $1$};
  \draw (0.008,0.25) -- (-0.008,0.25) node[left] {\scriptsize $\tfrac14$};
  % the crisp manifold: Var = mu(1-mu), where the crispness penalty is zero
  \draw[very thick,blue!55!black,domain=0:1,samples=140,smooth] plot (\x,{\x*(1-\x)});
  \node[blue!55!black] at (0.5,0.283) {\small crisp manifold\, $\mathbb{E}[v(1{-}v)]=0$};
  % a soft interior unit: its crispness penalty is the vertical gap up to the manifold
  \node[dot,fill=black] (soft) at (0.56,0.10) {};
  \node[anchor=north] at (0.56,0.083) {\scriptsize soft unit};
  \draw[->,very thick,orange!85!black] (0.56,0.11) -- (0.56,{0.56*(1-0.56)-0.004});
  \node[orange!75!black,anchor=west,align=left] at (0.578,0.170)
        {\scriptsize crispness\\[-1pt]\scriptsize pressure};
  \node[anchor=east,align=right] at (0.542,0.174)
        {\scriptsize penalty\\[-1pt]\scriptsize $\mu(1{-}\mu){-}\mathrm{Var}$};
  % variance floor
  \draw[dashed,red!70!black] (0,0.03) -- (1.06,0.03);
  \node[red!70!black,anchor=west] at (1.06,0.03) {\small $\mathrm{Var}=\tau$};
  % dead constants: the two endpoints of the manifold (Var = 0)
  \node[dot,fill=red!80!black] at (0,0) {};
  \node[dot,fill=red!80!black] at (1,0) {};
  \node[red!75!black,align=center,anchor=north] at (0.5,-0.028)
        {\small \textsc{constant} (dead): manifold endpoints, $\mathrm{Var}=0$ --- below the floor};
  % live selective detectors: interior of the manifold
  \node[dot,fill=green!45!black] at (0.30,{0.30*0.70}) {};
  \node[dot,fill=green!45!black] at (0.50,0.25) {};
  \node[green!40!black,anchor=west] at (0.185,0.135)
        {\small \textsc{selective} (live): $\mathrm{Var}\ge\tau$};
  \draw[green!40!black,thin] (0.30,0.150) -- (0.30,0.202); % leader to the green dot
\end{tikzpicture}

%% file: understand.tex
\section{Reading the Units}
\label{sec:understand}

\begin{figure}[htbp]
\centering
\includegraphics[width=\textwidth]{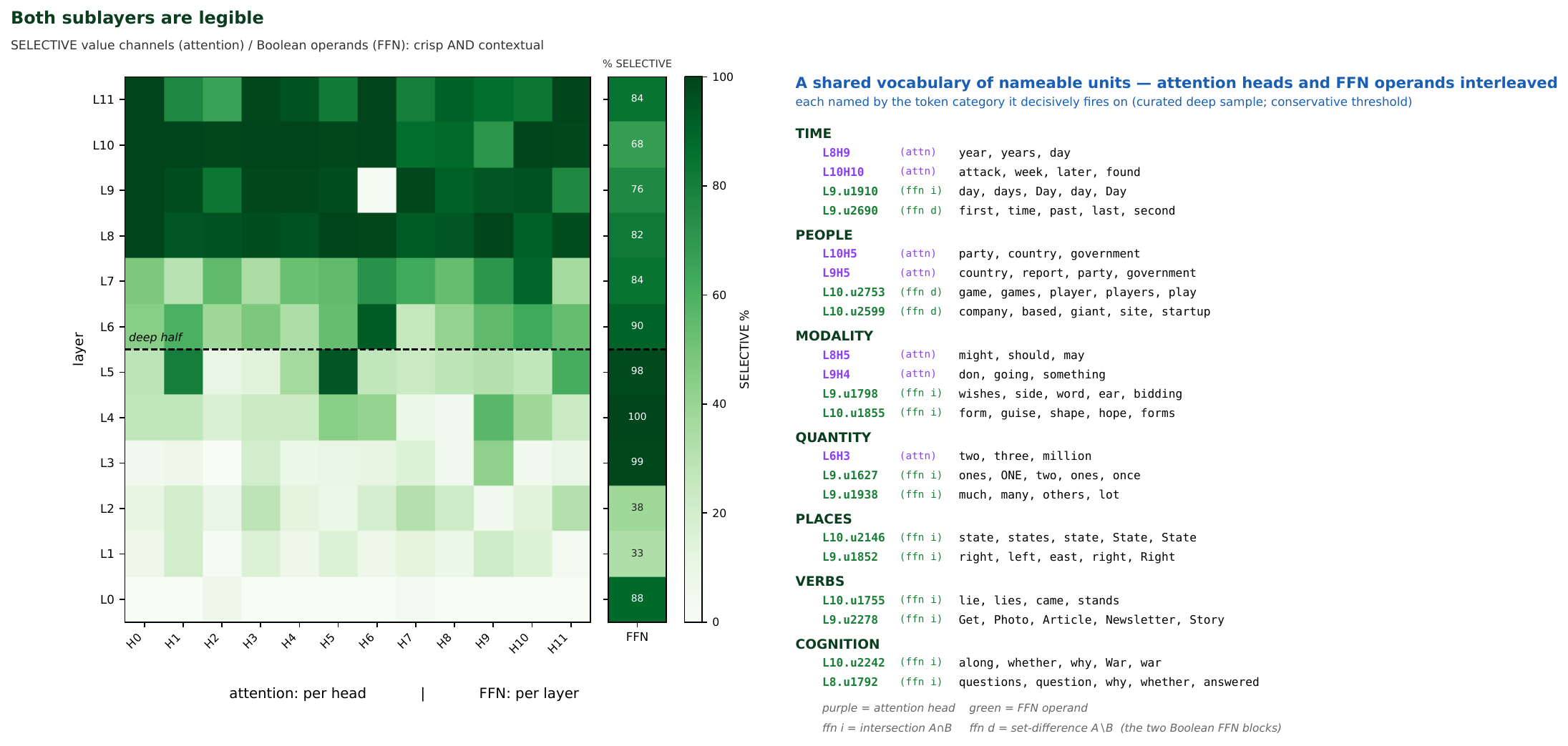}
\caption{Legibility of the annealed model on both sublayers. \emph{Left:} the fraction of \SELECTIVE{} (crisp \emph{and} contextual) channels per attention head ($12\times12$ grid) and per feed-forward layer (the adjacent strip, on the same colour scale; each strip cell is annotated with that layer's \SELECTIVE{} percentage). Attention legibility rises with depth --- shallow (L0--L5) $18\%$ against deep (L6--L11) $78\%$ --- while feed-forward legibility is high across most of the stack ($\ge82\%$ for L3--L8) and dips only at the earliest two layers. \emph{Right:} a curated sample of nameable deep units, attention heads (purple) and feed-forward units (green) interleaved under shared categories, each labelled by the token category it decisively fires on. Feed-forward units are tagged by their Boolean block --- \texttt{ffn i} for the intersection $A\cap B$, \texttt{ffn d} for the set-difference $A\setminus B$. A channel is \SELECTIVE{} when its activation variance is at least $0.003$ and over half its mass lies within $0.1$ of a rail; category labels use each unit's decisively-fired tokens (attention) or its logit-lens-promoted tokens (feed-forward, \citealp{geva2022promoting}). The sample is curated to the strongest per category ($15$ of $72$ deep heads and ${\sim}7000$ feed-forward units are nameable by this seven-way dictionary); the feed-forward strip is per-layer, coarser than the per-head attention grid. The near-white deep head (L9H6, ${\sim}1.6\%$) is a genuine non-selective outlier among its highly selective neighbours, not missing data. The attention panel extends the per-head legibility map of \citet[Fig.~4]{oskin2026attention} to the model trained with our objective, to the feed-forward sublayer, and to a merged catalog.}
\label{fig:hero}
\end{figure}

A \SELECTIVE{} unit detects a clean, input-dependent bit; naming \emph{what} that bit is means reading what the unit writes. The two questions come apart. This section reads the units of the annealed model in the correct per-layer frame, separates detection from naming as a measured population split, and isolates the part of that split that survives a learned readout. The model is legible on both sublayers, and many of its deep units are nameable by the token category they fire on (Figure~\ref{fig:hero}). The attention half of Figure~\ref{fig:hero} follows the per-head legibility map of \citet[Fig.~4]{oskin2026attention}, but is new here in three ways: it is measured on the model our training objective produces rather than the attention-only model of that work, it is joined to the feed-forward sublayer on the same scale (the two were never shown together), and its catalog interleaves attention heads with feed-forward units. It reports the legibility our objective induces across the whole layer, not a recomputation of that result.

\subsection{The reading problem}
\label{sec:understand-lens}
Project a unit's write through the raw unembedding --- the logit lens \citep{nostalgebraist2020logitlens} --- and it is legible at the boundary layers but reads as noise mid-stack. This is the known failure of the logit lens, and its known cause: the residual stream has no privileged basis \citep{elhage2021framework,elhage2023privilegedbasis}, so a mid-stack representation lives in a rotated frame that the fixed vocabulary projection does not align with. A learned per-layer affine map --- the tuned lens \citep{belrose2023tunedlens} --- recovers the legibility by rotating each layer back into the vocabulary frame before decoding. Across the stack, raw mid-layer decoding accuracy sits near the floor ($\sim\!0.05$--$0.08$) while a single learned linear lens lifts it four- to sevenfold ($0.30$--$0.36$). This is the standard logit-lens and tuned-lens picture, used here as a tool.

\subsection{A worked example: best of times}
\label{sec:understand-bestoftimes}
\begin{figure}[htbp]
\centering
\includegraphics[width=\textwidth]{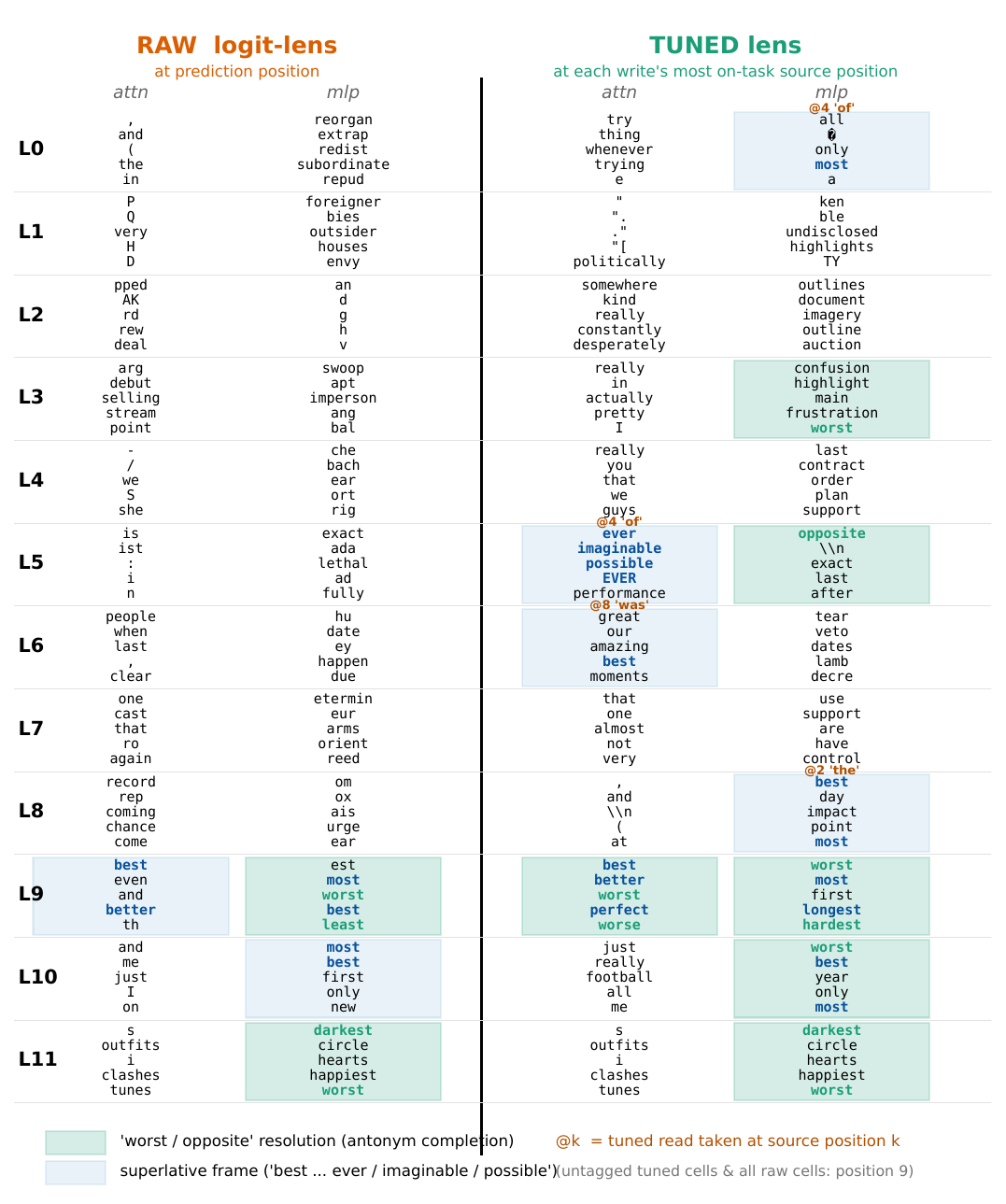}
\caption{The prompt \emph{``It was the best of times, it was the \_\_\_''} (completion: \emph{worst}), each layer's attention and feed-forward writes decoded under the raw logit lens (left) and the tuned lens \citep{belrose2023tunedlens} (right). The raw column reads at the final prediction position; the tuned column reads each write at its most on-task source position (tagged \texttt{@}$k$; untagged tuned cells and all raw cells are the final position). Raw: the \emph{best}$\to$\emph{worst} machinery is visible only at L9--L11. Tuned, six layers earlier: an L3 feed-forward unit promotes \emph{worst} and an L5 feed-forward unit the antonym \emph{opposite} (green, antonym completion), while the L5 attention head reads the superlative frame \emph{ever, imaginable, possible} at the \emph{best of times} position (blue). The middle stack computes the abstract structure; the deep layers (L9--L11) perform the lexical commit. The per-layer reading follows \citet{oskin2026boolglu}; the tuned-lens columns are new here, exposing the machinery six layers earlier than the raw logit lens.}
\label{fig:bestoftimes}
\end{figure}

The prompt \emph{``It was the best of times, it was the \_\_\_''} makes the frame concrete (Figure~\ref{fig:bestoftimes}). The model completes \emph{worst}. Tracing where a completion is assembled across a model's layers in this way follows the direct per-layer reading of \citet{oskin2026boolglu}; what is new here is reading it through the tuned lens rather than the raw logit lens, which moves the visible onset of the circuit six layers earlier and resolves which source position each layer draws from. Under the raw lens the \emph{best}$\to$\emph{worst} machinery appears only late, at layers $9$--$11$; everything earlier reads as noise. Under the tuned lens the same machinery is legible roughly six layers earlier, from L3: an L3 feed-forward unit already promotes \emph{worst}; an L5 feed-forward unit decodes to \emph{opposite}, a literal antonym operation; and the L5 attention head reads the superlative-completion frame \emph{ever, imaginable, possible} at the \emph{best of times} positions. The middle stack computes the abstract structure --- a superlative frame and an antonym operation --- and the deep layers do the lexical commit, with \emph{worst} crystallizing to rank one only in the final layers (L9--L11). The deep layers are already on task under the raw lens; the tuned lens helps only the rotated middle, which is exactly where the abstract computation lives.

\subsection{Detection is not naming}
\label{sec:understand-split}
\begin{figure}[htbp]
\centering
\includegraphics[width=0.8\textwidth]{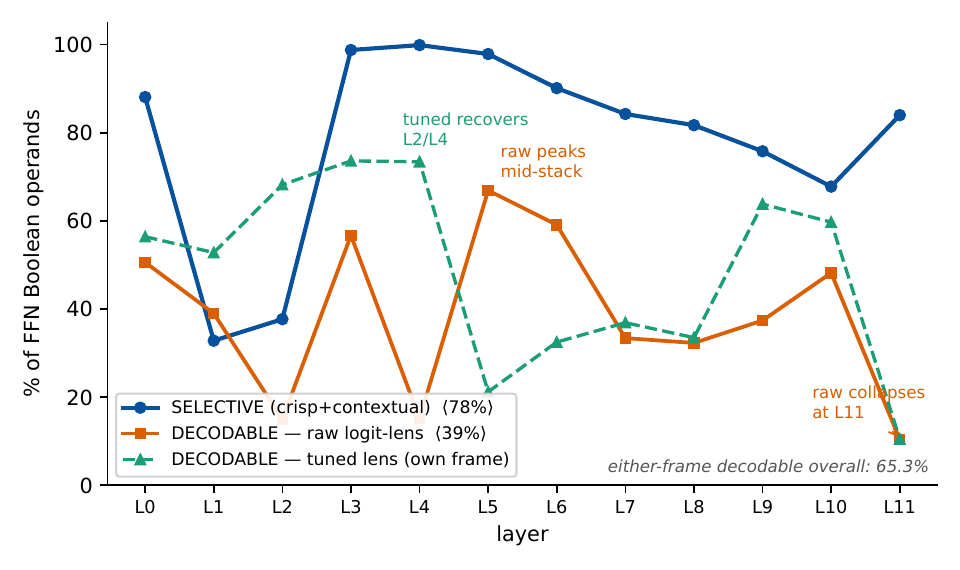}
\caption{Per-layer legibility of the annealed model's feed-forward operands: the fraction that are \SELECTIVE{} (detection-side) against the fraction \DECODABLE{} through the raw logit lens and through each unit's own tuned frame. Detection stays high across depth ($78\%$ overall); raw decodability peaks mid-stack and collapses at the final layer; the tuned frame recovers much of the mid-stack gap (either-frame $65.3\%$ overall). The detected bit is legible far more often than it is nameable.}
\label{fig:seldec}
\end{figure}

The distinction the construction lets us draw is between two legibility axes. \SELECTIVE{} is a detection-side property: the unit responds to a clean, input-dependent bit, measured on its bounded operands and independent of any readout basis. \DECODABLE{} is a promote-side property: what the unit \emph{writes} projects onto a nameable region of the vocabulary, in the sense of \citet{geva2022promoting}. On the annealed model these do not coincide (Figure~\ref{fig:seldec}). Of the feed-forward operands, $78\%$ are \SELECTIVE{} (at a stable crisp fraction of $0.90$; the count is reported at convergence, Section~\ref{sec:arch-metric}), but only $38.7\%$ ($7131$ of $18432$) decode through the logit lens to a nameable content category. On the attention side $50\%$ of value operands are \SELECTIVE{}, but only $18$ of $72$ deep heads read cleanly (all in L6--L11); of these, $15$ are nameable under the seven-way category scheme catalogued in Figure~\ref{fig:hero}. \DECODABLE{} is a proper subset of \SELECTIVE{}: selectivity is necessary but not sufficient for a name. The model detects cleanly but names only about half of what it detects, which quantifies the earlier observation that naming an operand's meaning is the harder frontier \citep{oskin2026attention}.

\subsection{Rotation, and the residue that survives it}
\label{sec:understand-control}
Half of the detected bits going unnamed by the raw lens invites the obvious reading: this is just logit-lens weakness, and a learned lens would name them. Re-reading each unit in its own tuned frame tests exactly that. Decodability rises from $38.7\%$ to $65.3\%$ when a unit is allowed to decode in either the raw or its own learned frame, and the recovery is concentrated where the rotation account predicts --- in the middle stack, where individual layers jump sharply (for example one layer moving from $15\%$ to $73\%$). Rotation therefore explains a large share of the gap: much of what looked undecodable was decodable in the wrong basis.

The remaining part does not close. About $35\%$ of the units resist even the learned lens. The residue is not uniform. Layers within the stack carry heterogeneous per-unit frames rather than one shared rotation, so a lens fit to the residual bulk helps bulk-frame units and can hurt units that were already canonical --- consistent with the absence of a privileged basis rather than a single recoverable rotation. And the final layer's low decodability ($\sim\!10\%$) is genuine, not rotation: its lens is near the identity, and the units there write fine-grained, non-token-peaked logit adjustments. Two properties are what remain after the known recovery is subtracted: this supra-lexical residue, and the layerwise \emph{lexicalization trajectory} --- a concept is non-lexical mid-stack and becomes token-decodable only in the final one or two layers. The closest prior framings are function directions that are steerable without being decodable \citep{nadaf2026steerable} and the polytope reading of features as globally distributed but locally aligned \citep{black2022polytope}; neither states the trajectory on units that are legible by construction.

The tuned-lens recovery is a general property of transformers \citep{belrose2023tunedlens} and applies as well to a \GELU{} model. What is specific here is the split between detection and naming, and the trajectory along which naming appears, measured on units whose detection side is legible by construction.

\subsection{Where a concept lives: redundant tokens, distributed concepts}
\label{sec:understand-diffuse}
The supra-lexical residue raises a question the construction is unusually well placed to answer: when the model computes a fuzzy conjunction, does the computation \emph{live} on a unit? Because each intersection unit is a named operator $A\cap B$, we can ablate one and re-run. Zeroing any single intersection unit's write moves the loss by essentially nothing (median $\Delta \approx 0$), yet zeroing the whole intersection block across the stack costs $0.65$ nats. The load is real but no single unit carries it. That is consistent with two very different pictures --- the same conjunction computed \emph{redundantly} across many copies, or one concept computed as a \emph{distributed} code with no single-unit home --- and single-unit ablation cannot tell them apart. Rank can.

\begin{figure}[htbp]
\centering
\includegraphics[width=\textwidth]{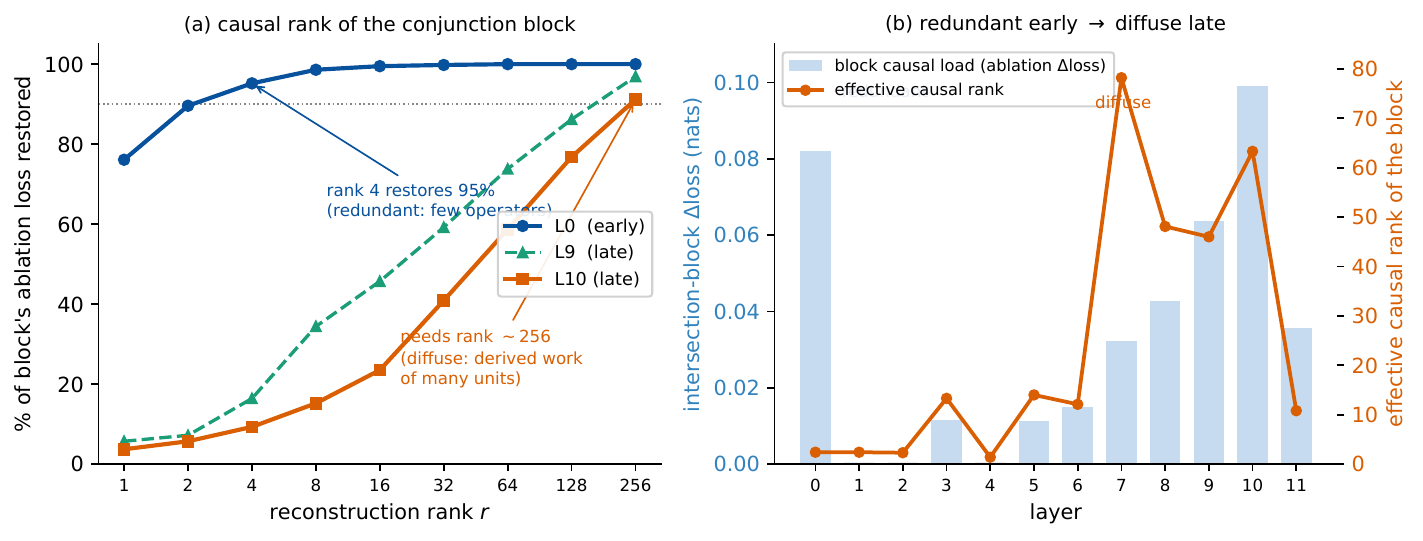}
\caption{How the fuzzy-conjunction block computes concepts, redundant early and distributed late. \emph{(a)} Ablate a layer's intersection block, add back a rank-$r$ reconstruction of its residual contribution, and sweep $r$: at L0 rank $4$ restores $95\%$ (a few operators, replicated), while at L9--L10 restoration needs rank ${\sim}256$ (a distributed code). \emph{(b)} Per layer, the block's causal load (bars, ablation $\Delta$loss) and its effective causal rank (line). The rank rises with depth as the computation moves from redundant to distributed. Single-unit ablation is near-zero everywhere, so neither panel is visible from ablating one unit.}
\label{fig:diffuse}
\end{figure}

We separate them causally (Figure~\ref{fig:diffuse}a). For a layer's intersection block we take its per-token residual contribution, ablate the block, and add back only a rank-$r$ reconstruction of that contribution, sweeping $r$. If the block computes a few operators replicated many times, a low-rank reconstruction restores the loss; if it is a distributed code, the loss returns only as $r$ grows. The two regimes are stark and depth-ordered. At layer $0$, rank $4$ restores $95\%$ of the block's effect: the block computes only a handful of distinct operators, replicated across the layer ($\sim\!180$ units per operator), each reproducible from a \emph{single} unit whose activation correlates above $0.9$ with the operator. At layers $9$--$10$, the same reconstruction needs rank ${\sim}256$: the computation is genuinely distributed, and the used operators are derived works of $15$--$40$ units that no single unit reproduces (best single-unit correlation $0.4$--$0.7$). The effective causal rank of the block climbs monotonically with depth while the redundancy factor falls from $183\times$ to $14\times$ (Figure~\ref{fig:diffuse}b). A single-unit read, which succeeds at the input, gives way to a distributed one at the output.

A second dissociation cuts across the first: the loudest structure is not the load-bearing one. At every layer a single dominant mode accounts for $74$--$90\%$ of the block's activation variance and is perfectly single-unit reproducible --- but at the deep layers it is causally almost inert (at L10 the rank-$1$ reconstruction restores only $3.7\%$ of the loss). The computation the model predicts with lives in the quiet, high-rank tail; a variance-ranked search for the important units finds exactly the wrong ones, which is why our first attempts to read these units by activation strength surfaced only a loud, meaningless mode.

What is distributed turns out to be a specific \emph{kind} of thing. We labelled tokens by five surface categories --- digits, punctuation, capitalized words, word-initial pieces, and newlines --- and asked, per category, whether a single intersection unit detects it as well as the whole block's best linear combination does. It does: across the deep layers the best single unit matches the full-block direction to within a few points of held-out \textsc{auc}, and a newline is detected by one unit at \textsc{auc} $1.00$ (Table~\ref{tab:tokenloc}). Token-level features are localized. The distributed operators are the other kind: reading the deep block's directions, the derived-work operators fire on abstract, cross-token structure and decode to no vocabulary token at all (write entropy at the vocabulary ceiling). The clearest one fires on clause and sentence boundaries --- \emph{But}, \emph{If}, \emph{When}, and reported-speech punctuation --- is spread over $21$ units with no single-unit reproducer, and carries a real if small causal load. The residue of Section~\ref{sec:understand-control} and the distribution here are the same phenomenon seen twice: the units the model keeps legible and single-homed are the token-shaped ones, and the concepts it distributes and leaves nameless are the supra-lexical ones.

\begin{table}[htbp]
\centering
\begin{tabular}{lcc}
\toprule
token feature & best single unit & whole block \\
\midrule
newline          & $1.00$ & $0.97$ \\
digit            & $0.97$ & $0.94$ \\
punctuation      & $0.86$ & $0.91$ \\
word-start piece & $0.81$ & $0.87$ \\
capitalized word & $0.75$ & $0.82$ \\
\bottomrule
\end{tabular}
\caption{Token-level features are localized to single units. Held-out detection \textsc{auc} (sign-corrected) at L10 --- the deep layer where abstract concepts are most distributed --- for the best individual intersection unit versus the whole block's best linear combination of units. A single unit is within a few points of the entire block; a newline is one unit. This is the opposite of the deep supra-lexical operators, which no single unit reproduces (Section~\ref{sec:understand-diffuse}).}
\label{tab:tokenloc}
\end{table}

This connects the construction to the superposition account of ordinary networks. Conventional transformers are argued to compute logical functions of features --- conjunctions included --- \emph{delocalized} in superposition, spread across neurons for efficiency \citep{elhage2022toy,hanni2024uand}, and such feature interactions are recoverable only post hoc \citep{marks2024sparse}. Making the operator legible by construction does not remove that pressure: the model still builds each conjunction redundantly, and reserves genuine distribution for the abstract concepts it has no token for. Legibility of the \emph{operator} is not localization of the \emph{concept}. The gap this leaves --- a distributed, nameless concept that we can point to only as a rank-$r$ direction --- invites a \emph{between}-unit training pressure aimed at it. Section~\ref{sec:dial} adds one and finds it reshapes the circuit rather than compacting it: a decorrelation penalty trades this distribution for surgically editable units without concentrating the concept. Whether a concept can instead be \emph{compacted} onto fewer units at no quality cost is a different question, which we leave open (Section~\ref{sec:discussion}).

%% file: edit.tex
\section{Editing the Units}
\label{sec:edit}

A legible unit should be a better place to make a change. If a coordinate carries a fixed, stated meaning, an edit aimed at that meaning should land there and nowhere else. We test this directly. The edit \emph{mechanism} is standard: a value-side promotion that adds $\beta\,E[T]$ to a unit's down-projection column, $W_o[:,u]\mathrel{+}=\beta\,E[T]$, the same operation used to inject a fact by writing into a feed-forward value \citep{geva2022promoting,dai2022knowledge,meng2022rome}. What is under test is not the mechanism but the \emph{site}: whether a legible-by-construction unit is a cleaner edit location than a conventional one. We measure a site by two quantities. \emph{Efficacy} is whether the promoted concept lands in the model's output where the unit fires; \emph{collateral} is $\mathrm{kl}_{\mathrm{off}}$, the KL divergence between the original and edited next-token distributions on the positions where the unit does \emph{not} fire --- the ROME notion of locality \citep{meng2022rome}. To compare sites fairly across bounded and unbounded units, we scale each unit's $\beta$ to a matched efficacy and read off the collateral it incurs to get there.

\subsection{Crisp firing predicts locality}
\label{sec:edit-crisp}
\begin{figure}[htbp]
\centering
\includegraphics[width=0.75\textwidth]{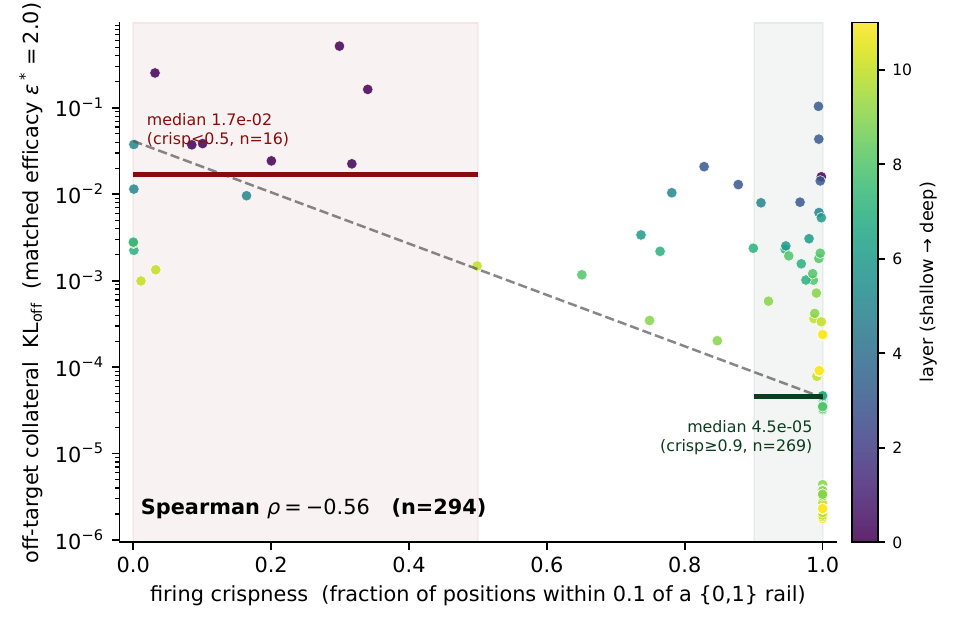}
\caption{Firing crispness predicts edit locality, within the annealed model. Each point is one Boolean feed-forward edit-site unit ($N{=}294$): its off-target collateral $\mathrm{kl}_{\mathrm{off}}$ at matched efficacy ($\mathrm{eff}^\ast=2.0$) against its firing crispness (fraction of positions within $0.1$ of a rail), coloured by layer. Crisper units leak less (Spearman $\rho=-0.56$); the crisp ($\ge0.9$) units carry a median collateral of $4.5\times10^{-5}$ against $1.7\times10^{-2}$ for the diffuse ($<0.5$) units, a factor of ${\sim}370$. The diffuse, high-collateral corner is a small group of shallow (mostly L0) units, and the crisp, low-collateral units are the deep ones --- which is why the edit advantage grows with depth.}
\label{fig:editcrisp}
\end{figure}

Within the showcase (\emph{anneal}) model, how crisply a unit fires predicts how local its edit stays (Figure~\ref{fig:editcrisp}). Across feed-forward units the rank correlation between firing crispness and $\log\mathrm{kl}_{\mathrm{off}}$ is $-0.56$: the crisper the detector, the less it leaks. The extremes are far apart --- units with firing crispness $\ge0.9$ carry a median off-target collateral of $\mathrm{kl}_{\mathrm{off}}=4.5\times10^{-5}$, against $1.7\times10^{-2}$ for units below $0.5$, a factor of roughly $370$. Firing density predicts collateral in the same direction ($+0.43$): a unit active on more positions leaks on more. A crisp, sparse detector is a small, well-insulated switch; a diffuse one spreads its edit across everything it touches.

\subsection{Boolean units are more targeted than \GELU{} units}
\label{sec:edit-vsgelu}
\begin{figure}[htbp]
\centering
\includegraphics[width=0.8\textwidth]{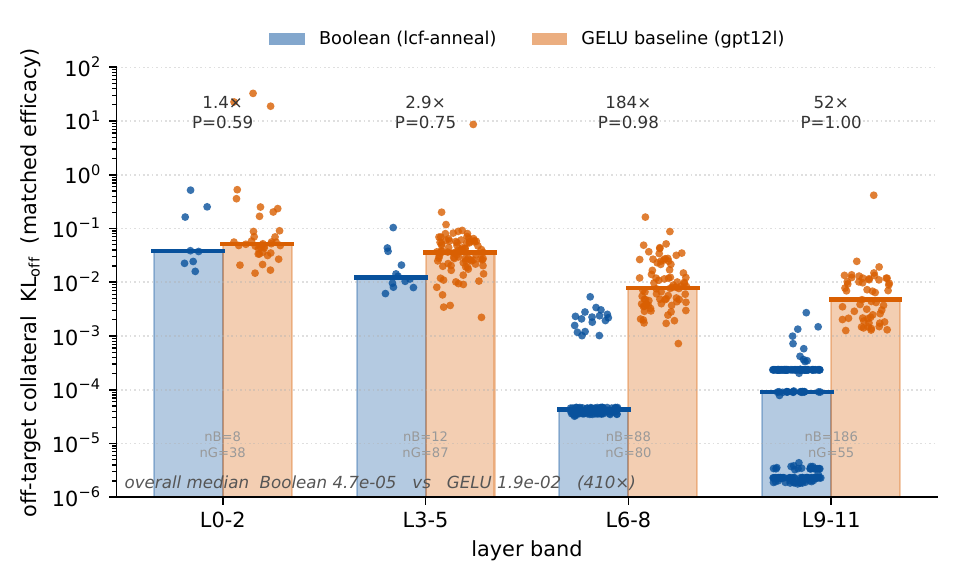}
\caption{Off-target collateral $\mathrm{kl}_{\mathrm{off}}$ per layer band at matched efficacy ($\mathrm{eff}^\ast=2.0$ nats), Boolean units (\emph{anneal}) against the conventional \GELU{} baseline; the count of Boolean and \GELU{} units ($n_B/n_G$) is printed under each band. The targetedness advantage emerges with depth: Boolean units are no better than \GELU{} in the earliest layers (L0--L2, $1.4\times$), where their outputs barely fire, but $184\times$ more local at L6--L8 and $52\times$ at L9--L11, where the edit sites concentrate (win probability $0.98$--$1.00$). Even where the earliest-layer operands are selective, the Boolean products ($A\cap B$, $A\setminus B$) they form rarely fire; layers L1, L2, and L4 have no editable Boolean output unit at all.}
\label{fig:kloff}
\end{figure}

The advantage holds between models, but it emerges with depth. Against the conventional \GELU{} baseline of Section~\ref{sec:arch-setup}, at matched efficacy the Boolean feed-forward units are no better targeted than \GELU{} in the earliest layers, where their outputs barely fire (L0--L2, $1.4\times$; L3--L5, $2.9\times$), and become far more local through the middle and deep stack, where the editable units concentrate (L6--L8, $184\times$; L9--L11, $52\times$; win probability $0.98$--$1.00$; Figure~\ref{fig:kloff}). The reason is the firing property of Section~\ref{sec:edit-crisp}: an edit stays local when its unit fires crisply and sparsely, and that firing is itself a deep-layer property, so the advantage appears where the firing does. Layers L1, L2, and L4 hold no editable Boolean output unit at all --- their set-operator products almost never rise above the firing threshold.

\subsection{A conjunctive edit}
\label{sec:edit-conj}
\begin{figure}[htbp]
\centering
\includegraphics[width=0.7\textwidth]{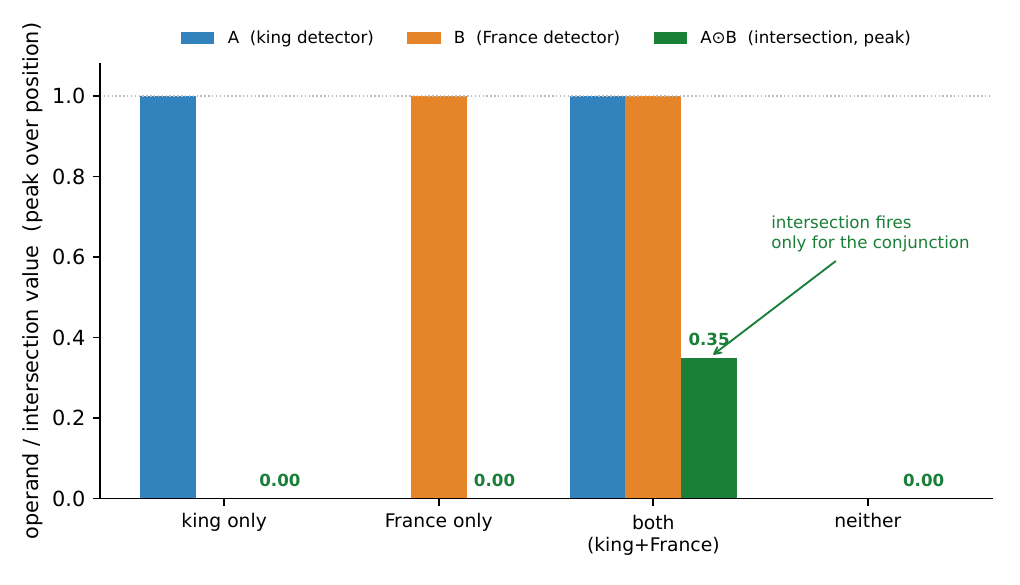}
\caption{A conjunctive $A\cap B$ edit. Two difference-of-means concept directions --- ``king'' and ``France'' --- are installed into the two operand banks of one Boolean feed-forward unit (layer $6$, $j{=}17$); the unit's unchanged fuzzy-AND $A\odot B$ then fires only for the conjunction. Across four context classes, $A$ fires on king contexts, $B$ on France, and the intersection (peak over positions) is nonzero \emph{only} for ``both'' ($0.35$), exactly $0$ for either concept alone and for neither. A single-preactivation \GELU{} unit cannot represent this. Verified on held-out contexts.}
\label{fig:conj}
\end{figure}

Legibility also buys an edit a conventional unit cannot express. This edit differs from the promotions above: rather than editing an \emph{existing} unit, it \emph{installs} a new detector into one, so the site is free and any set-operator coordinate serves; we use a single unit ($j{=}17$) at layer $6$. Into its two operand banks we install a selective direction for each of two concepts --- for a concept $C$, the difference of means $\mathrm{mean}(x\mid C\text{ in context})-\mathrm{mean}(x\mid\text{not})$ --- so that $A=\sig(W_a x)$ fires when ``king'' is present and $B=\sig(W_b x)$ when ``France'' is. The unit's own fuzzy-AND $A\odot B$ then computes the conjunction, unchanged. It gates cleanly (Figure~\ref{fig:conj}): $A$ fires only on king contexts, $B$ only on France, and the intersection is nonzero \emph{only} when both are present --- exactly zero for either concept alone and for neither. A conventional \GELU{} unit cannot be edited this way: it has a single pre-activation, so it can add two directions but cannot gate on their conjunction. The distinction is exactly the one the architecture makes available --- a unit that combines two named operands rather than one summed input --- and it separates our edit from additive compositional steering, which co-activates two concepts in parallel rather than gating on their conjunction \citep{turner2023actadd,ilharco2023task}.

Two caveats bound the demonstration. The installed direction is a lightweight data fit (a difference of means), not the raw token embedding: an embedding is shaped for the output geometry and is not selective read from a mid-layer residual --- the same read-frame effect of Section~\ref{sec:understand}. And the intersection's peak magnitude ($0.35$) is capped by single-position superposition, since two concepts rarely co-encode maximally on one token; the separation from the other classes (exactly $0$) is clean.

\subsection{The predictor is firing, not meaning}
\label{sec:edit-through}
What makes a good edit site is not what the interpretability reflex would suggest. The predictor of low collateral is \emph{bounded, crisp, sparse firing}. It is \emph{not} decodability: whether a unit's write decodes to a nameable token under a logit lens is, if anything, \emph{anti}-correlated with edit locality (rank $+0.48$ in the wrong direction), and it is not the learned crisp gate $\alpha$ either ($+0.15$). Locality is a structural consequence of the firing pattern, and that firing pattern is exactly what the selectivity objective of Section~\ref{sec:train} induces.

Knowing \emph{where} to edit is separate from knowing \emph{how}, and only one of the two is open. Where follows from the firing pattern: a unit fires sparsely and crisply, so an edit at it stays confined. How depends on the target. For a target that is a vocabulary token, we know how as well --- write its direction in the layer-native frame (Section~\ref{sec:edit-native}). What is open is the target that is a \emph{logical-operation output}: a set-operator unit's write $A\cap B$ or $A\setminus B$ combines the features its two operands detect, and those features are most likely concepts the model minted for its own prediction during training, not human categories with lexical names. There may be no word to look up for ``$A$ and-not $B$'' because there is no human concept there at all --- so steering a unit toward such an output means \emph{deriving} its direction from data, as the conjunctive edit fit king and France, rather than reading off a token. This is the detection-versus-naming gap of Section~\ref{sec:understand} on the editing side: the site is in hand, and for a named concept the write is too; naming the model's own invented logical outputs is the part with no dictionary.

The editability is the downstream payoff of the training objective: the mechanism remains ROME's, and what the objective supplies is the quality of the site.

This also decouples the result from the standard critique that interpretability is a weak proxy for steering utility \citep{wu2025interpretability}. That critique targets a pipeline in which features are \emph{discovered} post hoc, \emph{clamped} at inference, and judged by how decodable they are. Our units differ on all three axes: they are \emph{constructed} rather than discovered, \emph{edited in the weights} rather than clamped in activations, and scored by \emph{off-target} KL rather than by decodability --- which, as above, does not predict site quality at all.

\subsection{Deep edits need the layer-native frame}
\label{sec:edit-native}
Promoting a token bumps its logit; it does not make the model \emph{reason} as though the concept were present. The two are different edits and want different frames. A surface promotion writes the token embedding $E[T]$ and rides the additive residual highway: the write persists to the output essentially unchanged and raises $T$'s own logit, but injected late it has no machinery left to condition and diverges from any natural continuation. A deep behavioral edit --- one that makes the downstream computation behave as if $T$ were in the context --- instead needs the write the layer itself would have received, a layer-native vector $d_L(T)$. We obtain $d_L(T)$ empirically as a difference of means between contexts that contain the concept and matched contexts that do not; this construction is the diff-of-means of contrastive activation addition and activation steering, localized as in ROME \citep{rimsky2024caa,turner2023actadd,meng2022rome}. The mechanism is theirs; only the frame the write is made in is new.

The two frames behave very differently against the distribution a genuine mention of $T$ would produce (Table~\ref{tab:native}). The native write $d_L(T)$ reproduces that genuine-$T$-present next-token distribution, with a KL at or below the no-edit baseline: for ``France'' it reaches $0.45$ against a baseline of $0.77$, for ``ocean'' $0.31$ against $0.60$. The raw embedding write $E[T]$ only lifts $T$'s own logit, and injected late it diverges catastrophically, with KLs of $13$--$30$. Qualitatively the native write promotes the genuinely associated tokens --- Paris, French, Atlantic --- into a natural distribution, where the raw write spikes the concept word itself. Reading a unit and writing to it are dual operations with opposite correct frames: reading forward projects through the rotated per-layer frame, while a persistent additive write is made in the raw embedding frame, and a \emph{behavioral} write is made in the layer-native frame recovered by contrast.

\begin{table}[htbp]
\centering
\begin{tabular}{lccc}
\toprule
target $T$ & no-edit baseline & native $d_L(T)$ & raw $E[T]$ (late) \\
\midrule
France & 0.77 & \textbf{0.45} & 13--30 \\
ocean  & 0.60 & \textbf{0.31} & 13--30 \\
\bottomrule
\end{tabular}
\caption{Deep behavioral edit on the showcase model. A write is injected at a mid layer, and we report $\mathrm{KL}(P^\ast\,\|\,P_{\mathrm{edit}})$ between the edited model's next-token distribution $P_{\mathrm{edit}}$ and $P^\ast$, the distribution the \emph{unedited} model produces when a genuine mention of $T$ is prepended to the same context. \textbf{Lower is better}: a small KL means the edit makes the model continue \emph{as if} $T$ were actually present, rather than merely raising $T$'s own output logit. The \emph{no-edit baseline} column is the KL of the original prompt to $P^\ast$ --- how far the context already sits from genuine $T$-presence before any edit. The layer-native difference-of-means write $d_L(T)$ reaches \emph{at or below} that baseline, reproducing genuine presence --- and promoting the genuinely associated content (\emph{Paris}, \emph{French} for France) --- more faithfully than the unedited prompt itself. The raw embedding write $E[T]$ stays far from $P^\ast$ ($\mathrm{KL}$ $13$--$30$): it only spikes $T$'s own logit and, injected late, has no downstream layers left to turn that spike into genuine reasoning.}
\label{tab:native}
\end{table}

\subsection{What legibility does and does not buy}
\label{sec:edit-limits}
Legibility buys \emph{locality}, not \emph{magnitude}. Under a literal all-positions KL the Boolean and \GELU{} edits are within $1.4\times$ ($0.019$ against $0.026$): the legible edit concentrates its change on the concept rather than shrinking the total shift it makes, so its advantage is the off-target locality of Section~\ref{sec:edit-vsgelu}, not a smaller overall divergence. The edit operation is ROME's and the deep write is contrastive activation addition's; what is ours is the quality of the site, quantified by off-target KL on non-firing positions. These results are on single target tokens, and the conjunctive edit's magnitude is capped by lexical superposition at a single position. The nearest prior context is feature steering on sparse-autoencoder features, where more monosemantic units incur less collateral \citep{templeton2024scaling,neuronlens2025ranges}; we differ by constructing the units rather than discovering them, editing them in the weights rather than clamping them at inference, and in a quantified off-target locality.

%% file: dial.tex
\section{A Legibility Dial: Trading Fan-Out for Fan-In}
\label{sec:dial}

Section~\ref{sec:understand-diffuse} left the model computing each concept \emph{redundantly}: a fuzzy conjunction is produced across many copies, and read --- as we show below --- by many downstream units. That structure is legible in one sense and awkward in another. It is a \emph{fan-out} circuit: a few concepts, each broadcast to a large set of readers. Broadcasting is efficient (one concept, amortized), but it makes the concept hard to touch. To edit what a baseline concept means, one must alter every redundant copy \emph{and} accept that the change propagates to all of its readers. This section asks whether the circuit can be reshaped into something more editable, and what that costs. The intervention is a single training pressure, and it exposes a dial: at one end the reused, fan-out circuit of the preceding sections; at the other, a \emph{fan-in} circuit of independent units that are individually editable --- at no cost to quality.

\subsection{Reuse is correlation}
\label{sec:dial-idea}
The reason a concept is reused is also the reason it is hard to isolate. A building block that serves many computations --- a ``royalty'' feature feeding both \emph{king} and \emph{queen} contexts --- is, by definition, active whenever any of those computations is. Reuse across contexts \emph{is} co-occurrence, and co-occurrence is correlation. So a pressure that forbids units from correlating is, implicitly, a pressure against shared building blocks. We add one. Alongside the per-unit crispness and sparsity pressures of Section~\ref{sec:arch-pressure}, we introduce a \emph{between}-unit term that penalizes the mean squared off-diagonal correlation of the operand activations,
\begin{equation}
\mathcal{L}_{\mathrm{d}}=\lambda_{\mathrm{d}}\,\frac{1}{C(C-1)}\sum_{i\neq j} R_{ij}^{2},\qquad
R_{ij}=\frac{\mathrm{cov}(v_i,v_j)}{\sqrt{\mathrm{var}(v_i)\,\mathrm{var}(v_j)}},
\label{eq:diff}
\end{equation}
computed over each batch's tokens on the bounded operands $v=[A;B]$. The penalty is scale-free (the covariance normalization cancels the token count) and cheap (one $C\times C$ Gram matrix per block, no eigendecomposition). We train the annealed model of Section~\ref{sec:train} unchanged except for this one added term, sweeping $\lambda_{\mathrm{d}}$ over two orders of magnitude.

\subsection{The dial}
\label{sec:dial-sweep}
The penalty does exactly what it targets: mean off-diagonal operand correlation falls monotonically with $\lambda_{\mathrm{d}}$, from $0.51$ in the baseline to $0.016$ at the strongest setting (Table~\ref{tab:dial}). It costs nothing on quality. Development perplexity and \textsc{lambada} are flat across the sweep --- at the strongest setting the model is if anything marginally ahead ($0.239$ vs.\ $0.232$ \textsc{lambada} accuracy, $133$ vs.\ $143$ perplexity) --- so the intervention moves the circuit without moving the task. One thing does \emph{not} happen: the decorrelated model is not more legible per unit. Its fraction of \SELECTIVE{} detectors is \emph{lower} than the baseline's at convergence ($62\%$ vs.\ $78\%$); the operands stay crisp but a larger share drift below the contextuality floor. Per-unit legibility is not the axis the dial moves.\footnote{The \SELECTIVE{} \emph{count} is a knife-edge statistic here: the crisp population sits just above the variance floor $\tau$ (mean across-context variance ${\sim}3$--$5\tau$), so a small global drift in variance moves a large count across the threshold, and intermediate checkpoints can swing by tens of points. We therefore report converged (end-of-epoch) values and the stable continuous quantity (crisp fraction ${\approx}0.9$) alongside the count.}

What the dial moves is the \emph{shape} of the circuit, and it moves it from fan-out to fan-in. We measure both directions of wiring on the converged models. \emph{Fan-out} --- how many downstream units read a given concept direction --- collapses from a mean of $21$ (with hubs read by up to $1200$ units, most of a layer) to a mean of $1.4$ (largest hub $46$). \emph{Fan-in} --- how many atoms are needed to reconstruct a unit's input, its description length under a sparse decomposition over tokens, earlier concepts, and attention directions (the apparatus of Section~\ref{sec:understand})\footnote{This decomposition is orthogonal matching pursuit over a fixed interpretable dictionary, in the spirit of the sparse-coding interpretability line \citep{tamkin2024codebook}, applied to a unit's \emph{input}; it is apparatus, not a method contribution.} --- rises from $2$ atoms to $15$. The two numbers are the two faces of one change: without the pressure, a few concepts broadcast widely and each unit reads almost nothing (short description, high fan-out); with it, no concept is shared, so every unit must assemble its input from a wide, near-disjoint set (long description, low fan-out). The depth of the circuit does not change --- causal chains are one to two layers in both models (Section~\ref{sec:understand-diffuse}). It is a change in width, and in its direction. A single log-ratio captures it: $F=\log_{10}(\text{fan-in}/\text{fan-out})$ moves from $-0.98$ (broadcast) to $+1.02$ (aggregate) as the correlation falls, crossing zero --- equal parts fan-out and fan-in --- at $\lambda_{\mathrm{d}}\approx3\times10^{-2}$ (Table~\ref{tab:dial}), so the reshaping is graded by the dial, not a switch between two fixed regimes.

\begin{table}[htbp]
\centering
\begin{tabular}{lccccc}
\toprule
$\lambda_{\mathrm{d}}$ & off-diag $|R|$ & fan-in & fan-out & $F=\log_{10}\!\frac{\text{in}}{\text{out}}$ & \SELECTIVE{}\% \\
 & (target) & (atoms) & (readers) & & \\
\midrule
$0$ (baseline)      & $0.51$  & $2.2$  & $20.8$ & $-0.98$ & $78.7$ \\
$10^{-3}$           & $0.11$  & $1.2$  & $7.6$  & $-0.79$ & $71.2$ \\
$3{\times}10^{-3}$  & $0.09$  & $1.2$  & $7.7$  & $-0.81$ & $70.3$ \\
$3{\times}10^{-2}$  & $0.05$  & $5.1$  & $4.1$  & $+0.10$ & $63.4$ \\
$3{\times}10^{-1}$  & $0.03$  & $9.4$  & $2.7$  & $+0.54$ & $51.6$ \\
$1$                 & $0.016$ & $14.9$ & $1.4$  & $+1.02$ & $62.2$ \\
\bottomrule
\end{tabular}
\caption{The decorrelation dial, converged models (\SELECTIVE{}\% averaged over $200$k/$250$k/end-of-epoch to remove its knife-edge fluctuation; Section~\ref{sec:understand-diffuse} footnote). As $\lambda_{\mathrm{d}}$ rises, off-diagonal operand correlation --- the penalty target --- falls monotonically ($0.51\to0.016$), and the circuit shape follows: fan-in (atoms to reconstruct a unit's input) rises, fan-out (readers per concept) falls, and their log-ratio $F$ flips near-symmetrically from $-0.98$ (broadcast) to $+1.02$ (aggregate), crossing zero --- equal parts fan-out and fan-in --- at $\lambda_{\mathrm{d}}\approx3\times10^{-2}$, $|R|\approx0.05$. The one wrinkle is at the weakest settings, where decorrelation first makes units \emph{simpler} (fan-in dips to $1.2$) before the flip. Per-unit legibility does \emph{not} track the dial: \SELECTIVE{}\% falls and is genuinely non-monotone (intermediate $\lambda_{\mathrm{d}}$ least crisp, strongest re-crispens; reproduced across all three checkpoints, not a sampling artifact). The dial moves circuit \emph{shape}, not per-unit legibility.}
\label{tab:dial}
\end{table}

\subsection{Fan-in units are surgically editable}
\label{sec:dial-edit}
\begin{figure}[htbp]
\centering
\includegraphics[width=\textwidth]{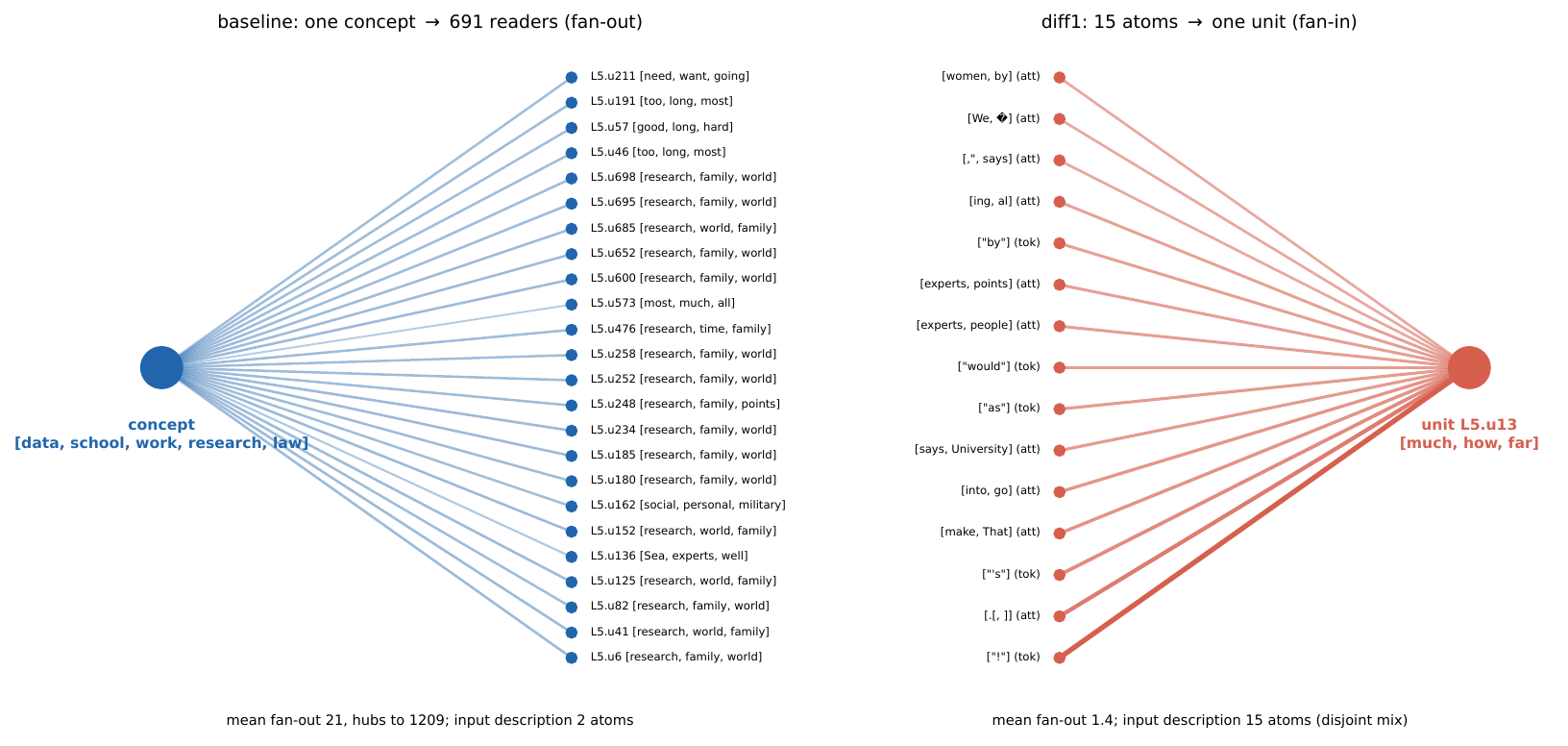}
\caption{Fan-out versus fan-in, from real circuits (edge width $\propto$ weight; nodes labelled by their most-fired tokens). \emph{Left:} a single baseline concept (firing on \emph{data, school, work, research, law} --- an institutions/information direction) and its actual reader units --- operand-read weights that align with it --- $24$ of its $691$ readers shown, ranked by content coherence. The readers are topic detectors, mostly near-duplicates firing on \emph{research, family, world} plus a distinct institutional-qualifier unit (\emph{social, personal, military}) --- one detector, largely replicated, the redundancy that fan-out entails. Without the pressure the model reuses a concept across a large reader set (population mean fan-out $21$/layer, hubs to ${\sim}1200$), so each unit's own input is short ($2$ atoms) but an edit ripples to every reader. \emph{Right:} a single decorrelated ($\lambda_{\mathrm{d}}{=}1$) unit and its actual input atoms --- the sparse decomposition of what it fires on over \{propagated tokens, earlier concepts, attention directions\}. With the pressure a unit aggregates ${\sim}15$ disjoint inputs (here a degree unit, \emph{how much}/\emph{how far}) and is itself read by almost no one (fan-out $1.4$), so editing the concept touches one unit rather than a $562$-unit group. Same shallow depth, opposite width, equal quality.}
\label{fig:dial}
\end{figure}

The point of the reshaping is that fan-in units can be edited where fan-out concepts cannot. Editability is governed by fan-\emph{out}: an edit to a unit propagates to everything that reads it, so a hub read by a thousand units cannot be changed without collateral. We make the comparison directly (Figure~\ref{fig:dial}). In the baseline, moving a concept requires editing a co-firing group of a median $562$ units --- the redundant copies --- and the edit is not local: at matched on-target effect, off-target \textsc{kl} \emph{grows} with edit strength ($0.002\to0.005\to0.014$ across efficacy bins), the signature of the hub rippling outward to its readers. In the decorrelated model the same concept lives on a single unit, and the edit stays put: off-target \textsc{kl} is flat and roughly two orders of magnitude smaller ($\sim\!10^{-4}$, independent of strength). A concept that was a $562$-unit rippling tangle becomes one addressable, surgically editable unit. This is the sense in which decorrelation buys editability: not by making units more readable, but by removing the reuse that made them entangled.

\begin{figure}[htbp]
\centering
\includegraphics[width=\textwidth]{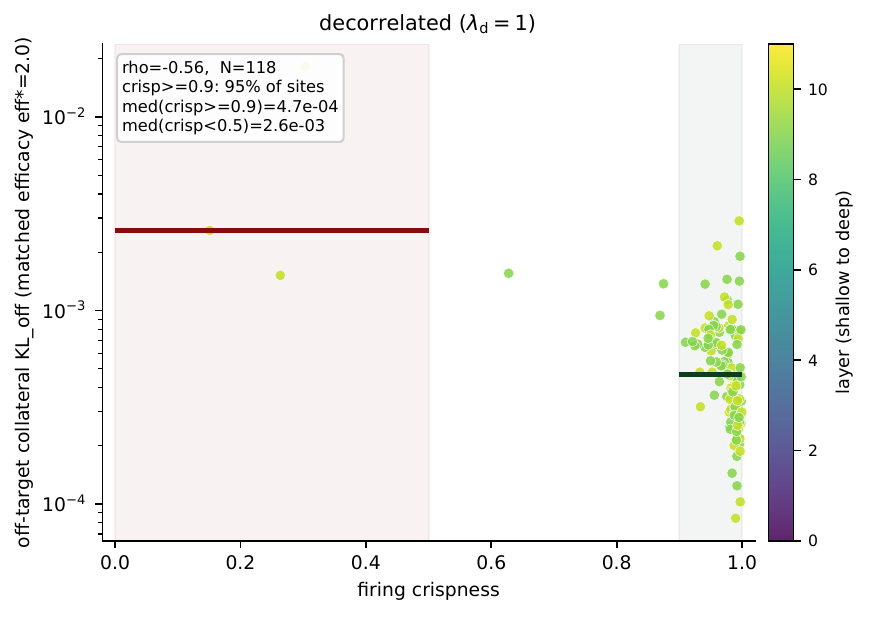}
\caption{Figure~\ref{fig:editcrisp}'s analysis on the decorrelated model ($\lambda_{\mathrm{d}}{=}1$): firing crispness versus off-target collateral $\mathrm{kl}_{\mathrm{off}}$ at matched efficacy, coloured by layer. The editable units are almost all crisp ($95\%$) and deep, banded around $5\times10^{-4}$, with the leaky diffuse corner that Figure~\ref{fig:editcrisp} shows for the baseline (shallow units at crisp$<0.5$, median $1.7\times10^{-2}$) nearly empty. Decorrelation removes the diffuse edit sites rather than improving the best ones.}
\label{fig:editcrispcontrast}
\end{figure}

The reshaping also homogenizes the edit sites. Section~\ref{sec:edit-crisp} found, within the baseline, that how crisply a unit fires predicts how local its edit stays (Figure~\ref{fig:editcrisp}): edit sites span the crispness range, and the high-collateral, leaky corner is a group of shallow, diffuse units. Running the identical analysis on the decorrelated model, that spread collapses (Figure~\ref{fig:editcrispcontrast}). Its editable Boolean units are almost all crisp ($95\%$ fire above $0.9$, against a broader baseline spread) and confined to the deep layers, and their off-target collateral falls in a narrow band around $5\times10^{-4}$: the leaky diffuse corner is essentially empty. The decorrelated model does not reach the very lowest per-unit collateral of the baseline's best crisp units, but it has no leaky sites at all --- nothing above ${\sim}3\times10^{-3}$, against a baseline tail that reaches $10^{-1}$. Because each concept now lives on a single unit rather than a redundant group, the population of edit sites is uniformly of the crisp, well-insulated kind, rather than a mixture that includes diffuse leakers.

\subsection{A named explanation of a prediction}
\label{sec:dial-explain}
\begin{figure}[tp]
\centering
\includegraphics[width=\textwidth]{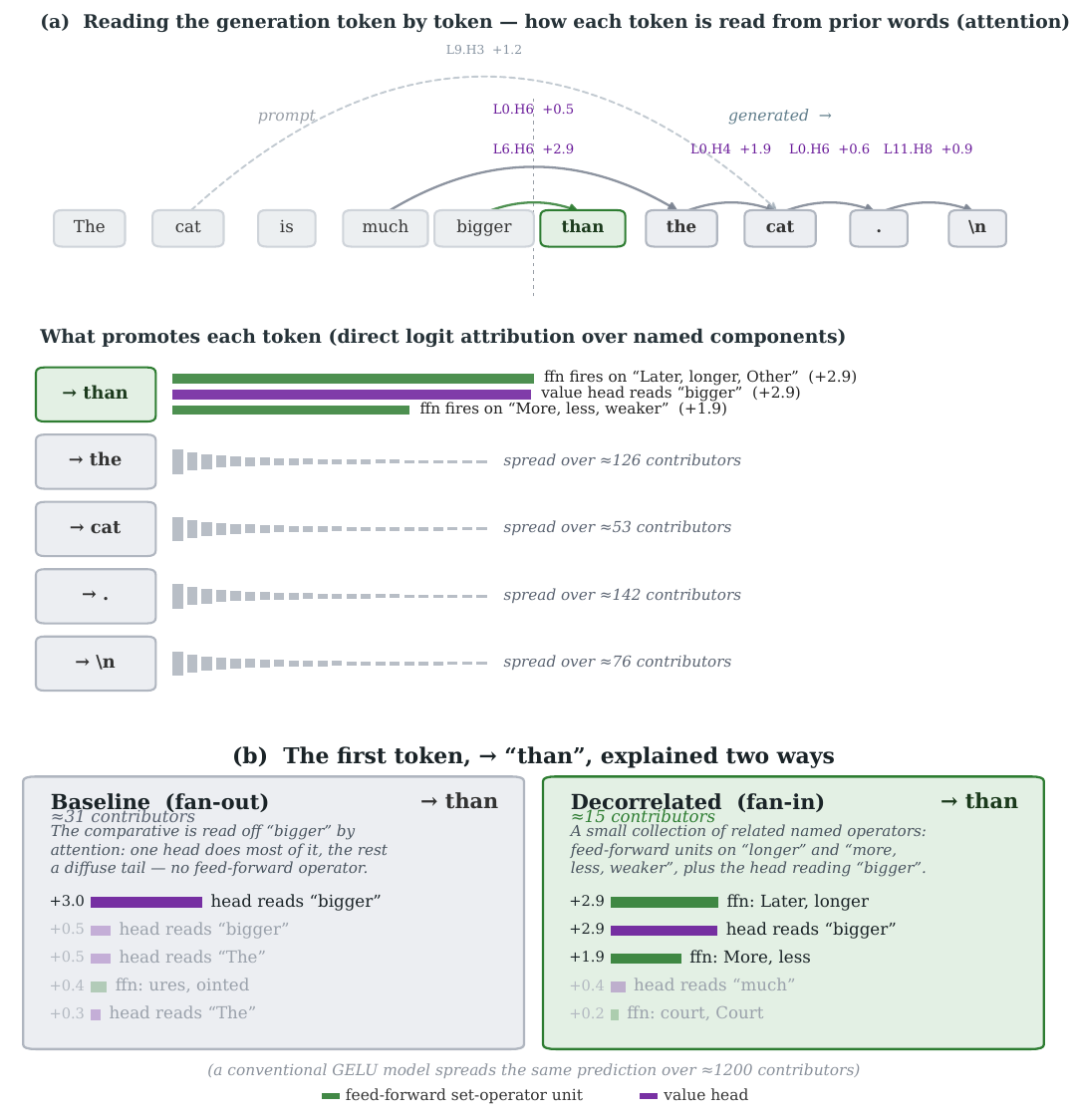}
\caption{\textbf{Reading a prediction, and what decorrelation does to it.} \emph{(a)} The decorrelated model generating from ``\emph{The cat is much bigger}'' token by token: each produced token is arced back to the prior word its top value head reads (a lighter dashed arc marks a strong secondary head --- here the induction head \texttt{L9.H3} that copies the earlier \emph{cat} to predict \emph{cat}); attention is legible at every step, and below, its direct-logit attribution over named components --- a short list of named operations where the attribution concentrates (the feed-forward-driven \emph{than}), a diffuse spread otherwise (function words and punctuation). \emph{(b)} A zoom on the first token, \emph{than}, explained by the baseline (fan-out) and decorrelated (fan-in) models over named components (green: feed-forward set-operator units; purple: value heads). Baseline: the comparative is read off \emph{bigger} by attention --- one value head carries most of it $(+3.0)$, the remainder a diffuse tail (${\approx}31$ effective contributors) --- with no feed-forward comparative operator among the top. Decorrelated: a small collection of related named operators carries it --- feed-forward units firing on \emph{longer} and on \emph{more, less, weaker}, plus the head reading \emph{bigger} (${\approx}15$). Bars show the five strongest contributors per model; a conventional \GELU{} model spreads the same prediction over ${\approx}1200$.}
\label{fig:namedcontrast}
\end{figure}

The end point of legibility is to explain a prediction. Because every component writes into the residual stream linearly, a predicted token's logit can be attributed to individual components by direct logit attribution. The point of the end-to-end construction is that \emph{both} halves that attribution lands on are legible: the feed-forward contributors are named set-operator units, and the attention contributors are the bounded value heads of \citet{oskin2026attention}, each read as a detector with a stated target. Every large contributor is therefore a named operation, and --- the property that makes it a sentence --- there are \emph{few} of them. This is the by-construction advantage, and it is one of \emph{sparsity}, not of nameability. Running the identical attribution on a conventional \GELU{} transformer of the same size, the neurons are still labelable by their fired tokens post hoc, and some are recognizably comparative (\emph{more, longer, larger}); what changes is that the prediction is \emph{diffuse}. The same \emph{than} is spread over ${\sim}1200$ effective contributors --- a thousand small named pieces no short list captures --- against ${\sim}15$ for the bounded model. The construction does not make units nameable where a conventional neuron is not; it makes the attribution sparse enough to state. This is the superposition that motivates the whole line \citep{elhage2022toy}, now measured on a single prediction: the standard model distributes the computation across a cloud of polysemantic units --- in both its feed-forward and its attention halves --- and what the bounded construction supplies is not names but concentration onto a legible few.

For the comparative continuation ``\emph{The cat is much bigger}'' the model predicts \emph{than}, and the attribution reads directly: \emph{than} is promoted by a feed-forward unit that fires on \emph{longer, more} $(+2.9)$, an attention head reading the comparative adjective \emph{bigger} $(+2.9)$, and a second feed-forward unit that fires on \emph{more, less, weaker} $(+1.9)$. The same three components carry \emph{than} across every comparative prompt we tried --- ``taller than his brother but shorter,'' ``the results were more significant'' --- with the head re-reading whichever comparative word is present (\emph{shorter}, \emph{more}). One can state, in words the reader already understands, why the token was produced.

Two things bound the result, and both echo the earlier findings. First, the explanation concentrates only for feed-forward-driven predictions: on the comparative tokens the effective number of contributors is ${\sim}15$, whereas on attention-driven or diffuse continuations (predicting \emph{not}, or \emph{be} after ``\emph{They will never}'') the attribution spreads over hundreds of components with no short named list --- the ceiling of confident, feed-forward-driven tokens already noted for these models \citep{oskin2026attention}. Second, the decorrelation sharpens the account rather than making it legible: the baseline explanation is already named (its top contributors are the same legible value heads, reading \emph{more}), but it is carried by heads alone with no feed-forward operator surfacing. The decorrelated model instead carries it on named comparative operators directly, and few of them (Figure~\ref{fig:namedcontrast}). Legibility of both sublayers supplies the named nodes; the dial makes the named explanation short and mechanistic.

\subsection{What flatness is worth}
\label{sec:dial-value}
The dial is a Pareto choice, not a free win. The fan-in model is the more editable and plausibly the more robust: with no load-bearing hubs, damage --- from pruning, quantization, or an edit --- stays local rather than propagating through a thousand dependents, and unlearning a capability need not disturb a shared abstraction. These are exactly the properties an inspection- or editing-first deployment wants, and the construction reaches them natively, without a post-hoc dictionary. What flatness spends is compression: reuse amortizes a concept across its readers, and the fan-in model rebuilds each one disjointly, at a longer description length --- a cost we did not see at $125$M parameters and one epoch but that should surface under tighter parameter budgets. The deeper open question is compositional: shared abstractions are plausibly what let a model recombine known concepts on novel inputs, and a circuit that reuses nothing may generalize worse out of distribution even at equal in-distribution perplexity. Our quality metrics are in-distribution and do not stress this; a compositional or out-of-distribution probe is the test that would tell us whether the fan-in end pays a generalization cost, and it is the natural next measurement.

This also sharpens where the construction sits relative to the decorrelation literature. Feature decorrelation and orthogonality regularizers have been proposed as routes to \emph{monosemanticity} --- more single-meaning, individually interpretable units \citep{elhage2022toy}. Our result is consistent with the letter of that and complicates its spirit: decorrelation does yield independent, individually addressable units, but at the cost of the shared structure that composition is built from, and \emph{not} at a gain in the per-unit \SELECTIVE{} count. What the pressure buys is not more legible atoms but a more \emph{editable} arrangement of them --- fewer readers per concept, at the price of a longer description per unit. Legibility, on this evidence, is not one axis but several, and they do not move together.

None of the structure in this section is special to the legible construction. A conventional transformer plausibly wires its concepts the same way --- reuse is efficient wherever a feature serves many computations --- and a decorrelation penalty on ordinary hidden activations, the objectives from which ours descends, would reshape fan-out into fan-in the same way. Every measurement here --- fan-out, description length, edit locality --- is defined on residual-stream activations and weights and runs unchanged on a $\GELU$ transformer. What the legible construction adds is not the structure but the ability to \emph{read} it: to name the concept a hub broadcasts (\emph{data, school, work, research}) or the atoms a fan-in unit collects, and to read it off a few named components rather than the diffuse cloud of polysemantic neurons a conventional model spreads the same content across. The dial is a general fact about transformer circuits; legibility is what lets us watch it turn.

%% file: discussion.tex
\section{Discussion}
\label{sec:discussion}

The result that ties the three parts together is the objective. A legible-by-construction architecture is a substrate, not a guarantee: its operators are legible only if training keeps them crisp and alive, and the natural pressure toward crispness, left alone, destroys exactly the property it was meant to create. Naming why --- the $\mu(1-\mu)-\mathrm{var}$ identity --- turns a tuning failure into a structural one and points at the minimal fix, a variance floor that is the legibility metric written as a loss. With that fix the substrate becomes usable: the units are legible enough to read in the rotated frame, and crisp enough to edit surgically. This is the sense in which the paper is a capstone of the legible feed-forward \citep{oskin2026boolglu} and legible attention \citep{oskin2026attention} line rather than a third architecture --- it is the objective that makes both work. The same move retires a hyperparameter. Prior work reserved a fixed fraction of feed-forward units as conventional $\GELU$ for trainability; the learned per-unit gate turns that reservation into a decision the model makes, and given the freedom it keeps no unit as pure $\GELU$ and drives the load-bearing computation crisp. What was a hand-set partition becomes an outcome, and the outcome is that the model wants the legible operators.

On quality the legible model is at parity. Against a conventional model the legible one is even on perplexity and LAMBADA and slightly ahead on grammar, and no result here is a benchmark improvement. The win is internal: the variance floor converts a pressure that collapses a model and wrecks its quality into one that recovers quality to parity while reaching the highest legibility in the family. A contribution of this kind is an objective and a set of measurements, not a leaderboard.

The construction does not make the model compute differently from a conventional transformer: the structure the later sections read off --- where concepts live, how redundantly they are built, how they are wired and how a decorrelation pressure reshapes them --- is structure an ordinary network plausibly shares, and every measurement is defined on activations and weights a $\GELU$ model has too. What the construction offers is \emph{legibility} --- that this structure can be read, its units named, its edits placed, without a post-hoc dictionary. The objective is what earns it.

A single further knob makes that payoff concrete and exposes its price (Section~\ref{sec:dial}). Adding a \emph{between}-unit decorrelation pressure, on top of the per-unit crispness and variance-floor pressures, reshapes the circuit along one axis at no cost to quality: from \emph{fan-out} --- a few concepts each reused by up to a thousand downstream units --- toward \emph{fan-in} --- many independent units, each aggregating a wide, disjoint input. It is a graded Pareto trade. Fan-in units are individually, surgically editable: a concept that lived in a $562$-unit redundant tangle becomes one addressable unit, edited with two orders of magnitude less off-target collateral. And a prediction's direct-logit attribution collapses from the diffuse cloud a conventional model spreads it across (${\sim}1200$ components) onto a short list of named operations (${\sim}15$) one can read as a sentence --- the by-construction advantage being sparsity of the attribution, not nameability, since both sublayers are already legible. What the trade spends is the compression reuse provides, whose cost on compositional generalization we do not yet measure.

The conjunctive edit points at a distinct payoff. Transformers form conjunctive and negated concepts --- ``this and that,'' ``this but not that'' --- as a matter of course, but conventionally in superposition, with nowhere single to address. Building the set operations in changes where such a concept can live: the architecture exposes a two-operand unit whose slots are addressable, so a conjunction can be \emph{written} at a specific \SELECTIVE{} site --- which is what the conjunctive edit does, installing a clean gate from two derived operand directions. Pointing to where a logical concept forms and writing to it is something a post-hoc method does not readily offer. Reading the model's \emph{own} conjunctions back out is a separate question, and the answer depends on how it builds them (Section~\ref{sec:understand-diffuse}): the concepts it computes \emph{redundantly}, a few operators replicated across a layer as at the input, are recoverable from any single copy, whereas the ones it \emph{distributes} across many units it leaves nameless --- recoverable only as a rank-$r$ direction, not from any single unit --- and it is these abstract, tokenless concepts that it spreads ever wider with depth. The localize-and-recover story is real but bounded: it holds where the model builds redundantly, not where it spreads.

Several limitations bound the contribution. All numbers are single seed at this scale; the direction is consistent across the ablations, but a headline of the form ``method A beats method B'' would need seeds we do not report. Legibility is free only up to a partition: past roughly half-Boolean the feed-forward layer meets a late-training instability, so the fully legible layer is a target the pressures approach rather than reach. And legibility is not the same as nameability. The units detect cleanly far more often than they can be named --- a concept is carried non-lexically through the middle of the network, becomes token-decodable only in the last layers, and may in the end be one the model minted for its own prediction rather than a human category with a name --- and closing that naming gap, rather than the detection gap the objective already handles, is where the interpretability work remains. On the editing side the mechanism is borrowed entirely; what the objective contributes is the quality of the site, and the gain there is locality rather than magnitude.

The naming gap also marks the next question. That a concept is legible on the detection side but non-lexical in mid-stack suggests the information a layer passes forward is neither token-shaped nor confined to a single unit, and reading it directly --- rather than through a detection metric or a per-unit lens --- would require decoding the residual stream in its own rotated, distributed basis. Section~\ref{sec:understand-diffuse} sharpens what stands in the way: token-level features are already single-homed, but the model distributes its abstract, nameless concepts across many units, and that distribution grows with depth. A concrete step is a pressure that \emph{concentrates} such a concept onto fewer units. Section~\ref{sec:dial} tried the between-unit pressure closest to hand --- decorrelation --- and found it does the reverse, spreading each concept wider in exchange for editability; the pressure this calls for is its dual, one that makes a \emph{concept} compact rather than a \emph{unit} decisive, aimed squarely at those distributed operators. Whether a concept can be compacted without costing quality is the natural test, and we leave it to future work.

%% file: conclusion.tex
\section{Conclusion}
\label{sec:conclusion}

A transformer built from bounded, named operators is a legible substrate, but a legible \emph{model} takes an objective. The crispness pressure legibility requires collapses its own operators into dead constants; the $\mu(1-\mu)-\mathrm{var}$ identity explains why, and a variance floor --- the legibility metric written as a loss --- repairs it, recovering quality while a learned per-unit gate retires the reserved-$\GELU$ partition and leaves the whole layer legible, on both sublayers at once. That model can then be read and acted on directly. In each layer's rotated frame its units separate a clean detection from a harder naming, detecting far more than they name and distributing exactly the concepts they cannot; being crisp and sparse, they are surgical edit sites and admit conjunctive edits no single neuron can express; and a between-unit decorrelation pressure exposes a dial from reused, fan-out circuits to independent, fan-in ones, trading compositional reuse for units editable one at a time and predictions readable as a short list of named operations. Quality holds at parity with a conventional model throughout. The objective is the contribution, and readability is what it buys --- a transformer whose computation can be read, and changed, not after the fact but from its own weights.

%% file: appendix.tex
\section{Trainability ceiling}
\label{sec:appendix-trainability}
The named fraction of the feed-forward layer is not a free hyperparameter. Below roughly half-Boolean the models train cleanly for the full epoch; above it they diverge late in training, and sooner the larger the Boolean fraction --- a fully-Boolean layer destabilizes early, a quarter-$\GELU$ layer much later, each with a sudden spike in instantaneous perplexity rather than a gradual drift. A linear-bypass path and an output normalization each delay the onset but neither prevents it, which points to a gradient pathology of the saturating operators rather than an unbounded-write problem \citep{oskin2026boolglu}. The consequence for this paper is stated in the main text: legibility is free at parity only up to a partition of the layer, and the pressures of Section~\ref{sec:train} operate below that ceiling.

\section{Hyperparameters}
\label{sec:appendix-hyper}
All models are $125$M-parameter, $12$-layer, $768$-wide transformers with learned positional embeddings, trained for one epoch on the same open-web corpus and tokenizer with an identical optimizer and schedule, and with \texttt{torch.compile} enabled. The selectivity pressures use $\lambda_{\mathrm{s}}=10^{-3}$ (sparsity) and $\lambda_{\mathrm{c}}=3\times10^{-3}$ (crispness). The variance floor uses $\lambda_{\mathrm{ctx}}=3\times10^{-3}$ and $\tau=0.003$, applied from the first step on both sublayers for the \emph{selective} model. The learned crisp fraction initializes $\alpha=\sig(\theta)$ at $\alpha\approx0.9$ (crisp-first), with an output-norm-weighted $\GELU$ tax $\lambda(1-\alpha)\lVert W_o[:,u]\rVert$; the \emph{anneal} schedule ramps $\lambda$ from $10^{-2}$ to $10^{-1}$ over training. The end-to-end models place every attention value under the Boolean form (Eq.~\ref{eq:boolean}).

\section{Probe methodology}
\label{sec:appendix-probe}
Legibility statistics are computed on held-out text. For a bounded operand channel with per-token values of mean $\mu_c$ and variance $\mathrm{var}_c$ over the sample, the channel is \emph{crisp} when the fraction of values within $0.1$ of a rail $\{0,1\}$ exceeds one half and \emph{contextual} when $\mathrm{var}_c\ge\tau=0.003$; \SELECTIVE{} is the conjunction, \CONSTANT{} is crisp with $\mathrm{var}_c<\tau$. \DECODABLE{} is measured by projecting a unit's write through the token unembedding and testing whether the top of the resulting distribution forms a nameable content category, following \citet{geva2022promoting}. The rotated-frame re-read fits a per-layer affine tuned lens \citep{belrose2023tunedlens} mapping each layer's residual to the pre-unembedding frame (ridge regression, with the final-layer map verified near the identity), and reports a unit as recovered when it decodes in the raw frame or its own tuned frame. The best-of-times attribution reads each layer's feed-forward and attention writes for the single prompt through both the raw and tuned lenses.